\documentclass[10pt,journal,compsoc]{IEEEtran}
\ifCLASSOPTIONcompsoc
  \usepackage[nocompress]{cite}
\else
  \usepackage{cite}
\fi
\usepackage{multirow}
\ifCLASSINFOpdf
\usepackage[pdftex]{graphicx}
\else
\usepackage[dvips]{graphicx}
\fi
\usepackage{amsmath}
\usepackage{amsfonts}
\usepackage{booktabs}
\usepackage{url}  
\usepackage{diagbox} 
\usepackage{tabularx}
\usepackage{times}
\usepackage{epsfig}
\usepackage{amssymb}
\usepackage{verbatim}
\usepackage{mathrsfs}
\usepackage{algorithm}
\usepackage{algpseudocode} 
\usepackage{multirow}
\usepackage{makecell}
\usepackage{bm}

\usepackage{amsthm}

\usepackage{color}

\ifCLASSOPTIONcompsoc
\usepackage[caption=false,font=normalsize,labelfont=sf,textfont=sf]{subfig}
\else
\usepackage[caption=false,font=footnotesize]{subfig}
\fi

\usepackage[pagebackref=false,breaklinks=true,colorlinks,bookmarks=false]{hyperref}

\begin{document}

\title{Contrastive Bayesian Analysis for\\Deep Metric Learning}

\author{Shichao~Kan,
        Zhiquan~He,
        Yigang~Cen,
        Yang~Li, 
        Vladimir~Mladenovic,
        and Zhihai~He
\IEEEcompsocitemizethanks{\IEEEcompsocthanksitem Corresponding authors: Yigang Cen and Zhihai He.
\IEEEcompsocthanksitem Shichao Kan is with the School of Computer Science and Engineering, Central South University, 410083, Changsha, Hunan, China, and also with the Institute of Information Science, School of Computer and Information Technology, Beijing Jiaotong University, Beijing 100044, China, and the Beijing Key Laboratory of Advanced Information Science and Network Technology, Beijing 100044, China (e-mail: kanshichao@csu.edu.cn).\protect
\IEEEcompsocthanksitem  Zhiquan He is with Guangdong Multimedia Information Service Engineering Technology Research Center, Shenzhen University, China, 518060 (e-mail:zhiquan@szu.edu.cn).\protect
\IEEEcompsocthanksitem  Yigang Cen is with the 
Institute of Information Science, School of Computer and Information Technology, Beijing Jiaotong University, Beijing 100044, China, and also with the Beijing Key Laboratory of Advanced Information Science and Network Technology, Beijing 100044, China (e-mail: ygcen@bjtu.edu.cn).\protect
\IEEEcompsocthanksitem  Yang Li is with the Department of Electrical Engineering and Computer Science, University of Missouri, Columbia, MO 65211, USA (e-mail: yltb5@mail.missouri.edu).\protect
\IEEEcompsocthanksitem Vladimir Mladenovic is with the Faculty of Technical Sciences University of Kragujevac, Cacak, Serbia (e-mail: vladimir.mladenovic@ftn.kg.ac.rs).\protect
\IEEEcompsocthanksitem Zhihai He is with the Department of Electrical and Electronic Engineering, Southern University of Science and Technology, Shenzhen, China, and also with the Pengcheng Lab, Shenzhen 518066, China (e-mail: hezh@sustech.edu.cn).\protect
\IEEEcompsocthanksitem Code project: https://github.com/kanshichao/CBML.\protect
}
}

\markboth{IEEE TRANSACTIONS ON PATTERN ANALYSIS AND MACHINE INTELLIGENCE}
{Kan \MakeLowercase{\textit{et al.}}: Contrastive Bayesian Analysis for Deep Metric Learning}

\IEEEtitleabstractindextext{
\begin{abstract}
Recent methods for deep metric learning have been focusing on designing different contrastive loss functions between positive and negative pairs of samples so that the learned feature embedding is able to pull positive samples of the same class closer and push negative samples from different classes away from each other. In this work, we recognize that there is a significant semantic gap between features at the intermediate feature layer and class labels at the final output layer. To bridge this gap, we develop a contrastive Bayesian analysis to characterize and model the posterior probabilities of image labels conditioned by their features similarity in a contrastive learning setting. This contrastive Bayesian analysis leads to a new loss function for deep metric learning. To improve the generalization capability of the proposed method onto new classes, we further extend the contrastive Bayesian loss with a metric variance constraint.
Our experimental results and ablation studies demonstrate that the proposed contrastive Bayesian metric learning method significantly improves the performance of deep metric learning in both supervised and pseudo-supervised scenarios, outperforming existing methods by a large margin. 
\end{abstract}

}

\maketitle
\IEEEdisplaynontitleabstractindextext
\IEEEpeerreviewmaketitle

\section{Introduction}
\label{se:section}

\IEEEPARstart{O}{ne} central task in computer vision and machine learning is to  learn and  generate features to characterize or represent images \cite{arXiv-He19}. 
As an important requirement, these features should be discriminative. Images with the same semantic labels should have similar features being aggregated into compact clusters in the high-dimensional feature space. Meanwhile, images from different classes should be well separated from each other. During the past a few years,  methods based on deep neural networks have made remarkable progress on learning discriminative features for images \cite{CVPR-HeZRS16}.

\begin{figure*}[t]
    \begin{center}
    \includegraphics[width=0.9\linewidth]{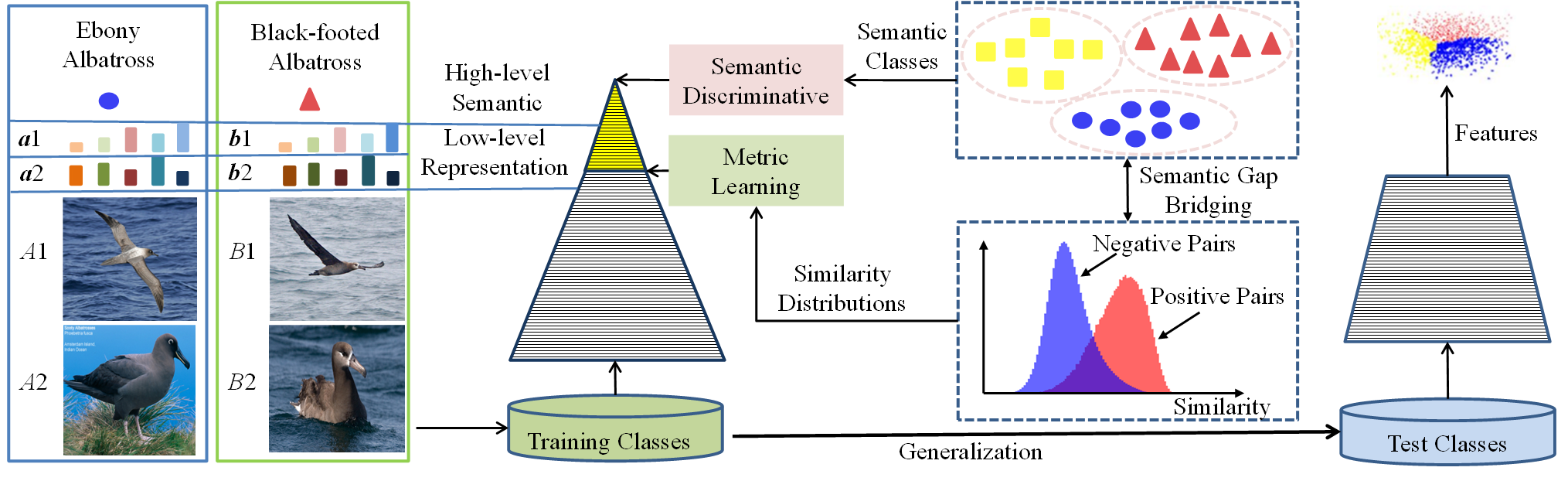}
    \end{center}
    \vspace{-2mm}
	\caption{Representation learning based on semantic supervision and metric learning on training classes and generalize to test classes. Because there is a semantic gap between high-level semantic and low-level representation in both inter-class and intra-class, for example, images $A1$ and $B1$ are from two different classes, but they have very similar features. Images $A1$ and $A2$ are from the same classes, but their features $\bm{a}1$ and $\bm{a}2$ are quite different. We aim to bridge this gap by modeling the relationship between semantic classes and similarity distributions based on contrastive Bayesian analysis (CBA).}
	\label{fig:problem}
	\vspace{-2mm}
\end{figure*}

Recently, research on deep metric learning or feature embedding \cite{CVPR-ChenD19,TIP-KanCHZZW19,PR-KanZHCCZ20,ECCV-KimGCLK18,PAMI-Opitz18} has achieved remarkable progress in image retrieval \cite{CVPR-LiuLQWT16,CVPR-SongXJS16}, fine-grained object classification and matching \cite{ICCVW-Krause0DF13,CIT-wah2011}, person re-identification \cite{Multimedia-Zhang20,zhang-local-2022}, and vehicle re-identification \cite{CVPR-LiuTWPH16}. 
Existing state-of-the-art methods  have been focusing on learning deep neural networks with carefully designed loss functions  to generate  discriminative features with the goal to minimize intra-class sample distance and maximize inter-class sample distance. For example, the contrastive loss \cite{CVPR-HadsellCL06}
captures the similarity between pairs of images from the same class and dissimilarity between samples from different classes.
The triplet loss \cite{CVPR-SchroffKP15} considers a triplet of samples with an anchor sample, one positive sample, and one negative sample. 
The purpose of triplet loss is to learn a distance metric by
which the anchor point is closer to the positive sample than the negative sample by a large margin. More recent deep metric learning methods, for example, lifted structured loss \cite{CVPR-SongXJS16}, proxy loss \cite{ICCV-Movshovitz-Attias17} and ranked list loss \cite{CVPR-WangHKHGR19}, further extend these loss functions by considering richer sample structure information. Some recent methods combine multiple loss functions and jointly optimize metric and softmax loss \cite{TIP-KanCHZZW19,CVPR-ZhangZLZ16}.  
Algorithms have also been developed to systematically discover or mine hard negative or positive samples \cite{CVPR-SuhHKL19,CVPR-ZhengCL019,CVPR-WangHHDS19}.
As pointed out in \cite{ICCV-Movshovitz-Attias17}, these loss functions heavily depend on how the  positive and negative samples are selected, which directly affects their metric learning performance and algorithm convergence rate. 

In this work, we observe that there are several major issues that have not been carefully addressed in existing deep metric learning. As illustrated in Fig. \ref{fig:problem}, the metric learning operates at the intermediate feature layer, aiming to preserve the label similarity relationships at the network output layer. This is a challenging task. For example, images $A1$ and $B1$ are from two different classes, but they have very similar features. Images $A1$ and $A2$ are from the same classes, but their features $\bm{a}1$ and $\bm{a}2$ are quite different. During metric learning, minimizing the feature distance between samples from the same class or maximizing their similarity does not necessarily guarantee that these samples can obtain similar representations. How do we analyze and model the inherent relationship between features and image labels and incorporate this analysis into the deep metric learning process emerges as an important research problem. 
Another important issue is the generalization problem. In many deep metric learning settings, the test classes are totally different from the training classes. We need to make sure that the features learned on the training classes can generalize well onto novel test classes. 

\begin{figure}[h]
    \begin{center}
    \includegraphics[width=1.0\linewidth]{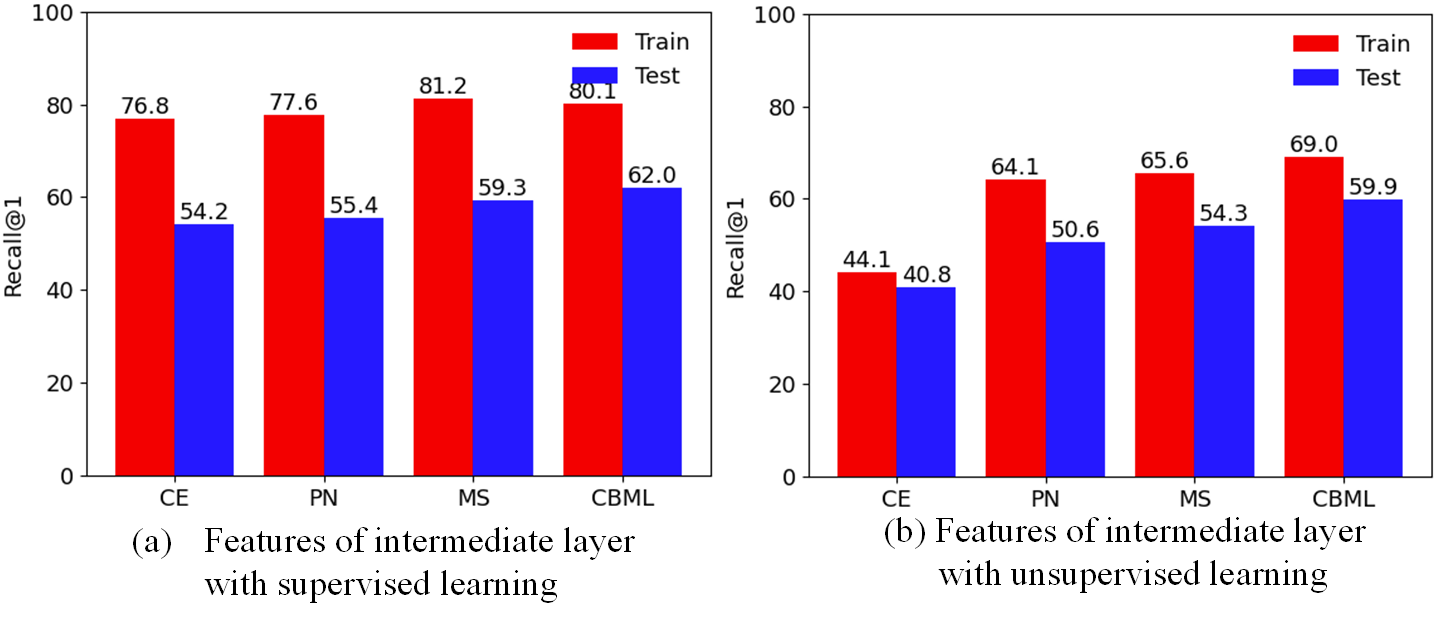}
    \end{center}
    \vspace{-4mm}
	\caption{ Retrieval performance of supervised (a) and unsupervised (b) training with cross entropy (CE) loss at the label layer and contrastive metric learning (i.e., PN, MS and CBML) at the intermediate layer on the CUB dataset based on the GoogLeNet backbone. The training and test classes are different. We compare the top-1 recall rates in an image retrieval setting. These models have been thoroughly trained separately.}
	\label{fig:supervised-gap}
	\vspace{-4mm}
\end{figure}

In the following experiment, we use an example to further demonstrate the importance of these two issues. As shown in Fig. \ref{fig:supervised-gap}, on the CUB dataset, we evaluate four different approaches for learning a deep neural network to extract image features. 
In the first approach, we directly train a GoogLeNet classification network based on the labeled training images using the cross entropy (CE) as the loss function. Once the network is fully trained, we use the output of the intermediate layer as the image feature (of size 1024). From the second to the fourth approaches, we use three metric learning methods, i.e., the proxynca (PN) loss \cite{ICCV-Movshovitz-Attias17}, the multi-similarity (MS) loss \cite{CVPR-WangHHDS19} and the contrastive Bayesian metric learning (CBML) loss to be developed in this paper, to optimize the GoogLeNet network, respectively. These metric learning methods are directly applied on the intermediate layer of the GoogLeNet. For these approaches, we test the learned features on both the training and test classes in an image retrieval setting. It should be pointed out that these models have been thoroughly trained separately. On the training set, we can see that the performance of the features optimized by the metric learning-based approaches is higher than that of the classification-based approach which uses the image labels to learn the image features. On the test set, the top-1 scores of the metric learning-based methods outperform the top-1 score of the classification-based approach. 
This experiment suggests that there is a significant difference between feature-level similarity and label-level semantic similarity. It also shows that our CBML method generalizes much better than other methods for deep metric learning.

In our proposed constrastive Bayesian analysis, we aim to address these two important issues. Specifically, we propose to analyze and model the inherent relationship between metric learning at the intermediate feature layer and their semantic labels at the final output layer based on a Bayesian conditional probability analysis. We develop this Bayesian analysis in a contrastive learning setting for positive and negative pairs and formulate a metric learning process. This new contrastive Bayesian analysis bridges the gap between the learned features of images and their class labels, resulting in a new loss function for deep metric learning.

Because the gap between the learned features of images and their class labels is bridged, the new loss function based on contrastive Bayesian analysis can easily overfit the training set which can result in performance degradation on novel classes. To improve the generalization capability of the proposed method onto new classes, we further extend the contrastive Bayesian loss with a metric variance constraint. 
Moreover, we couple this contrastive Bayesian analysis with clustering-based pseudo label generation in an iterative manner to achieve improved performance for pseudo-supervised deep metric learning. 
Our experimental results and ablation studies demonstrate that the proposed constrastive Bayesian metric learning (CBML) method improves the performance of deep metric learning, outperforming existing method by a large margin. 

The rest of this paper is organized as follows. Section 2 reviews the related work on deep metric learning. The proposed CBML method is presented in Section 3. Experimental results, performance comparison with state-of-the-art methods, and detailed ablation studies of our algorithm are provided in Section 4. Section 5 concludes the paper.

\section{Related Work and Major Contributions}
\label{sec:relatedwork}
This work is related to deep metric learning on both supervised and unsupervised scenarios. In this section, we review the existing methods on this topic and discuss the unique novelty and contributions of our proposed approach. 

\subsection{Supervised Deep Metric Learning}
Deep metric learning aims to learn useful semantic representations or feature embedding that can capture semantic similarity between data samples. One of the central tasks in deep metric learning is the design of metric loss functions. In the following, we will review sample-based, proxy-based and hybrid loss functions for deep metric learning, respectively.

\textbf{Sample-Based Loss Functions.}
Metric loss functions based on contrastive loss \cite{CVPR-HadsellCL06} and triplet loss \cite{CVPR-SchroffKP15} aim to minimize the distance between similar samples and maximize the distance between dissimilar ones. Sohn \textit{et al.} \cite{NIPS-Sohn16} improved the triplet loss and proposed an N-pair loss function for more robust feature embedding. Yu \textit{et al.} \cite{ICCV-Yu2019} proposed a tuplet margin loss based on randomly selected samples from each mini-batch. 
 Xun \textit{et al.} \cite{WACV-Xuan2020} proposed an easy positive triplet mining idea by mapping each training image to the most similar examples from the same class.
 Wang \textit{et al.} recognized that existing pairwise and triplet loss functions suffer from slow convergence due to a large proportion of trivial pairs or triplets as the model improves \cite{CVPR-WangHKHGR19}. To address this issue, they  proposed a simple and effective ranked list loss based on the rank structure of neighboring samples. 
Wang \textit{et al.} \cite{CVPR-WangHHDS19} proposed a multi-similarity loss based on a general pair weighting  scheme. Recently, they proposed a cross-batch memory \cite{CVPR-Wang20} mechanism that is able to  memorize the embeddings of past iterations to collect sufficient hard negative pairs across multiple mini-batches.
Song \textit{et al.} \cite{CVPR-SongXJS16} introduced a lifted structured feature embedding method based on a loss function of all positive pairs and negative pairs in the mini-batch. A cluster loss was introduced in \cite{CVPR-SongJR017} to optimize the clustering performance based on the normalized mutual information (NMI) metric \cite{Cambridge-Chri08}.  Ustinova \textit{et al.} \cite{NIPS-UstinovaL16} developed a histogram loss by estimating the distributions of matched and unmatched samples. Kumar \textit{et al.} \cite{CVPR-GC016} aimed to minimize the variance of distributions of matched and unmatched samples. Huang \textit{et al.} \cite{NIPS-HuangLT16} introduced a position-dependent deep metric (PDDM) method which adapts the metric function to the local feature structure so as to find high-quality hard samples \cite{ICCV-HarwoodGCRD17,CVPR-SuhHKL19,CVPR-ZhengCL019}.
Vasudeva \textit{et al.} \cite{Vasudeva-2021-ICCV}  proposed an optimal hard negatives (LoOp) mining method by considering the entire space between pairs of embeddings.
Cakir \textit{et al.} \cite{CVPR-Cakir0XKS19} developed a FastAP loss by optimizing the average precision. 
Roth \textit{et al.} \cite{ICCV-Roth19} studied the inter-class characteristics shared across object classes.  
Jacob \textit{et al.} \cite{ICCV-Jacob2019} proposed a high-order distribution-aware regularization scheme for deep metric learning. Chen \textit{et al.} \cite{NIPS-Chen19} proposed a curvilinear distance metric learning method that adaptively learns the nonlinear geometries of the training data. Deng \textit{et al.} \cite{CVPR-DengGXZ19} proposed an additive angular margin loss (ArcFace) for effective face recognition. Wen \textit{et al.} \cite{IJCV-WenZLQ19} developed a center loss for face recognition. Sun \textit{et al.} \cite{CVPR-Sun20} proposed a circle loss by re-weighting each similarity to highlight the less-optimized similarity scores.
Musgrave \textit{et al.} \cite{arxiv-Musgrave20} provided a comprehensive evaluation and comparison of different loss functions and found that these loss functions have similar performance.
To improve the generalization performance, Ko and Gu \cite{cvpr20-KoG20} proposed an embedding expansion method for metric learning losses, Venkataramanan \textit{et al.} \cite{ArXiv-abs-2106-04990} used mixup data augmentation approach to train a metric learning model.
Ko \textit{et al.} \cite{Ko-2021-ICCV} proposed a MemVir training strategy which stores both embedding features and class weights and treat them as additional virtual classes.

\textbf{Proxy-Based Loss Functions.} 
Proxies are a small set of representative samples inferred from the training set. The Proxy-NCA \cite{ICCV-Movshovitz-Attias17} is the first proxy-based loss function, which is realized by assigning a proxy for each class. This loss function minimizes the distance between the proxy and a positive example and maximizes the distance between the proxy and a negative example. Teh \textit{et al.} \cite{ECCV20-TehDT20} revisited ProxyNCA and proposed ProxyNCA++ by incorporating multiple improvements. Qian \textit{et al.} \cite{ICCV-Qian2019} introduced a soft triplet loss \cite{NIPS-Sohn16} using multiple proxies for each class to reflect intra-class variance. Aziere and Todorovic \cite{CVPR-AziereT19} proposed the manifold proxy loss to improve the embedding performance by extending the N-pair loss using proxies. By combining advantages of the sample-based and proxy-based methods, Kim \textit{et al.} \cite{CVPR20-KimKCK20} proposed the proxy anchor loss to boost the speed of convergence and improve the robustness of the learned embeddings. Gu \textit{et al.} \cite{AAAI-GuKK21} proposed a proxy synthesis method to generate synthetic embeddings and proxies to mimic unseen classes, which can improve the generalization capability of the embeddings.

\textbf{Hybrid Loss Functions.}
Incorporating more information into the feature embedding process to improve the robustness is another important research topic in deep metric learning. A number of methods based on attention modeling \cite{CVPR-ChenD19,ECCV-KimGCLK18,TIP-ZhouWMLGZ19}, ensemble learning \cite{CVPR-ChenD19,PAMI-Opitz18,CVPR-SanakoyeuTBO19,ECCV-XuanSP18}, and feature fusion  \cite{TIP-KanCHZZW19}
have been developed. These methods trained their models using multiple loss functions. 
Sanakoyeu \textit{et al.} \cite{CVPR-SanakoyeuTBO19} proposed an idea of learning separate distance metrics for different regions of the sample distribution using a divide and conquer approach. 
Chen \textit{et al.} \cite{CVPR-ChenD19}  proposed a hybrid-attention-based decoupling method by combining object attention and channel attention mechanisms. Also, they adopted the idea of feature ensemble with adversary learning for metric learning. Kim \textit{et al.} \cite{ECCV-KimGCLK18} proposed an attention-based ensemble with a multitask optimization model to generate robust feature embedding. Zhou \textit{et al.} \cite{TIP-ZhouWMLGZ19} introduced a foreground attention model with local regression and symmetric triplet loss functions for robust person re-identification. Opitz \textit{et al.} \cite{PAMI-Opitz18} developed a boosting ensemble strategy based on adversarial loss and triple loss functions to generate robust feature embeddings. Xuan \textit{et al.} \cite{ECCV-XuanSP18} proposed to learn multiple embedding functions and then combined them together to achieve robust feature embedding.
Kan \textit{et al.} \cite{TIP-KanCHZZW19} proposed a Fusion-Net model to fuse 4-RootHSV \cite{JVCIR-KanCCWVMZ17} features into GoogLeNet, and finally generated robust feature embedding by optimizing a multi-loss function. Recently, they proposed a LSCM-GNN method \cite{TIP-Kan22} to generate robust embedding by fusing K-NN embeddings based on a graph neural network. 

\subsection{Unsupervised Deep Metric Learning.}
Unsupervised deep metric learning is a more challenging task since the training classes have no labels and they does not overlap with the testing classes. One of the earliest works of unsupervised deep metric learning was directly using k-means clustering to assign pseudo-labels to features and updating the network parameters using pseudo-labels \cite{ECCV-CaronBJD18}. Latter, Iscen \textit{et al.} \cite{CVPR-IscenTAC18} adopted hard example mining based on manifold-aware to train feature embedding network. A momentum contrast (MoCo) method was proposed by He \textit{et al.} \cite{arXiv-He19} to realize visual representation learning in an unsupervised manner. A contrastive learning framework (simCLR) is proposed by Chen \textit{et al.} \cite{arXiv-Chen20} to realize effective unsupervised visual representation learning. Based on the positive augmentation invariant and negative separated properties, Ye \textit{et al.} \cite{CVPR-YeZYC19,PAMI-Ye20} proposed an instance method to optimize instance feature embedding. In order to generate more robust pseudo-labels for unsupervised deep metric learning,  Nguyen \textit{et al.} \cite{ArXiv/abs-2009-04091} proposed to use a deep clustering loss to learn centroids. Dutta \textit{et al.} \cite{ArXiv-abs-2008-09880,AAAI-DuttaHS20} proposed a graph-based clustering approach to generate labels. To approximate the positive concentrated and negative instance separated properties in the graph latent space, Ye \textit{et al.} \cite{CVPR-YeS20} proposed a probabilistic structural latent representation (PSLR) method. Recently, Kan \textit{et al.} \cite{CVPR-KanC0MH21} proposed a relative order analysis (ROA) and optimization method to optimize relative order of ranking examples for unsupervised deep metric learning. Li \textit{et al.} \cite{CVPR-0091KY0H21} proposed a spatial assembly networks (SAN) to realize effective supervised and unsupervised deep metric learning.

\subsection{Bayesian Analysis and Major Contributions.}
This work is also related to Bayesian analysis, which has been studied in metric learning.  Liong \textit{et al.} \cite{ECCV-LiongLG14} proposed a regularized Bayesian metric learning method to model and regulate the eigen-spectrums of within-class and between-class covariance matrices in a parametric manner, so that discriminative information can be better exploited for person re-identification.
A Bayesian active distance metric learning method was developed in \cite{arxiv-yang12} based on the variational method for classification application. 
A dynamic Bayesian metric learning model was developed in  \cite{CIKM-XiaoRMSL19} for personalized product search. 
Different from these works, we proposed to analyze and model the inherent relationship between sample labels and their similarity scores using a Bayesian conditional probability analysis approach for image retrieval. We also derive this new Bayesian analysis in a contrastive learning setting. 

Compared to existing work, the major contributions of this work can be summarized as follows.
(1) Existing methods on deep metric learning have been focusing on the contrastive metric loss function design at the intermediate feature layer. This work addresses the important limitation in existing approaches and bridges the semantic gap between features and image labels. We derive the constrastive Bayesian analysis to estimate the posterior probability of labels conditioned by their feature metric in a constrastive learning setting, which leads to a new loss function for deep metric learning.
(2) The second major contribution is that we extend the constrastive Bayesian analysis by considering the metric variance constraint and improve the generalization capability of the proposed method.
(3) Our new method based on contrastive Bayesian analysis has improved the performance of deep metric learning, outperforming existing state-of-the-art methods by a large margin. 

\section{Method}

\subsection{Problem Formulation}

Consider a set of images $\mathcal{I} = \{I_1, I_2, \cdots, I_n\}$ with labels $\mathcal{Y}=\{y_1,y_2,\cdots,y_n\}$.
These images are encoded into features
$\mathcal{X} = \{\bm{x}_1, \bm{x}_2, \cdots, \bm{x}_n\}$
where $\bm{x}_i = \mathcal{F}(I_i)$ and $\mathcal{F}(\cdot)$ is the encoder network.
Let $y_i=l_{\bm{x}_i}$ be the label for sample $\bm{x}_i$.
$m(\bm{x}_i,\bm{x}_j)$ represents the metric between features $\bm{x}_i$ and $\bm{x}_j$. 
For example, in our experiments, we use the cosine similarity as the metric at both the training stage and test stage.
Let $\mathcal{P}$ and $\mathcal{N}$ be the set of positive pairs and the set of negative pairs, respectively. Specifically, 
\begin{eqnarray}
  \mathcal{P}=\{(\bm{x}_i, \bm{x}_j)\ |\ l_{\bm{x}_i} =l_{\bm{x}_j}\}, \\
\mathcal{N}=\{(\bm{x}_i, \bm{x}_j)\ |\ l_{\bm{x}_i} \neq l_{\bm{x}_j}\}.
\end{eqnarray}
In existing methods for supervised metric learning, a typical approach is to learn a metric or similarity function $m(\bm{x}_i,\bm{x}_j)$ such that the similarity scores of positive pairs from $\mathcal{P}$ are maximized while the similarity scores of negative pairs from $\mathcal{N}$ are minimized.

As discussed in the previous section, we recognize that there is a semantic gap between features output from the the intermediate feature layer and actual sample labels. In other words, maximizing the similarity scores calculated based on features between samples in a class does not necessarily guarantee that the learned model can produce similar embeddings for these samples. To address this issue, we propose to analyze and characterize the relationship between sample labels and their similarity scores for positive and negative pairs and formulate metric learning process using a conditional probability analysis approach. 
Given the metric  $m(\bm{x}_i, \bm{x}_j)$, let
$p[l_{\bm{x}_i} \approx l_{\bm{x}_j}\ |\ m(\bm{x}_i, \bm{x}_j)]$
and $p[l_{\bm{x}_i} \not\approx l_{\bm{x}_j}\ |\ m(\bm{x}_i, \bm{x}_j)]$ be the posterior probabilities for the image pair $(\bm{x}_i, \bm{x}_j)$ to be in the positive pair set $\mathcal{P}$ and 
negative pair set $\mathcal{N}$, respectively. 
Note that a pair of image samples belongs to either $\mathcal{P}$ or $\mathcal{N}$.
We have 
\begin{equation}
    p[l_{\bm{x}_i} \approx l_{\bm{x}_j} \ |\ m(\bm{x}_i, \bm{x}_j)] +  p[l_{\bm{x}_i} \not\approx l_{\bm{x}_j} | m(\bm{x}_i, \bm{x}_j)] = 1.
\label{eq:pos-neg}
\end{equation}
During metric learning, for positive pairs in the training set
 $(\bm{x}_i,\bm{x}_j) \in \mathcal{P}$, we need to make sure that the learned feature representation is able to maximize the average probability of positive pairs conditioned by the metric function $m(\bm{x}_i, \bm{x}_j)$, in other words, to maximize $p[l_{\bm{x}_i} \approx l_{\bm{x}_j} | m(\bm{x}_i, \bm{x}_j)]$,  or to minimize the probability their conditional probability of becoming negative pairs, in other words, to minimize $p[l_{\bm{x}_i} \not\approx l_{\bm{x}_j} | m(\bm{x}_i, \bm{x}_j)]$.
 Similarly, for negative pairs $(\bm{x}_i,\bm{x}_j) \in \mathcal{N}$, we need to maximize the conditional probability  $p[l_{\bm{x}_i} \not\approx l_{\bm{x}_j} | m(\bm{x}_i, \bm{x}_j)]$,  or minimize the conditional probability  $p[l_{\bm{x}_i} \approx l_{\bm{x}_j} | m(\bm{x}_i, \bm{x}_j)]$.
 Based on this, we introduce the loss function for deep metric learning. 
 Specifically, for a given image sample $\bm{x}_i$, let
 \begin{equation}
     \mathcal{P}_i =\{\bm{x}_j | (\bm{x}_i, \bm{x}_j) \in \mathcal{P}\}, \quad
     \mathcal{N}_i =\{\bm{x}_j | (\bm{x}_i, \bm{x}_j) \in \mathcal{N}\}     
 \end{equation}
 be the set of positive and negative samples for $\bm{x}_i$, respectively.
 Our metric learning method aims to minimize the following objective function for positive and negative pairs:
 \begin{equation}
 \begin{split}
\mathcal{L}_0 & = -\mathbb{E}^A_{\bm{x}_i\in\mathcal{X}}\log\left\{\mathbb{E}^H_{\mathbf{x}_j\in\mathcal{P}_i}\left\{p[l_{\bm{x}_i} \approx l_{\bm{x}_j} | m(\bm{x}_i, \bm{x}_j)]\right\}\right\} \\
     & - \mathbb{E}^A_{\bm{x}_i\in\mathcal{X}}\log\left\{\mathbb{E}^H_{\mathbf{x}_j\in\mathcal{N}_i}\{p[l_{\bm{x}_i} \not\approx l_{\bm{x}_j} | m(\bm{x}_i, \bm{x}_j)]\}\right\}
\end{split}
\label{eq-org-loss}
 \end{equation}
 where $\mathbb{E}^A_{\bm{x}_i\in\mathcal{X}}\{\cdot \}$  represents the arithmetic average operation over the set $\mathcal{X}$.  
 $\mathbb{E}^H_{\bm{x}_i\in\mathcal{X}}\{\cdot \}$  represents the harmonic average operation. 
 $\log\{\cdot\}$ is the logarithmic function. 
  For a set of positive number $\{a_1, a_2, \cdots, a_N\}$, its harmonic average is defined as 
 \begin{equation}
     \mathbb{E}^H(a_1, a_2, \cdots, a_N) = \left[\frac{1}{N}\sum_{i=1}^N \frac{1}{a_i}\right]^{-1}.
    \label{eq-ha}
 \end{equation}
 It has been demonstrated that the harmonic average is always no greater than the arithmetic average
 \begin{equation}
    \mathbb{E}^H(a_1, a_2, \cdots, a_N) =  \left[\frac{1}{N}\sum_{i=1}^N \frac{1}{a_i}\right]^{-1} \le \frac{1}{N}\sum_{i=1}^{N}a_i.
 \end{equation}
 In (\ref{eq-org-loss}), we choose to use the harmonic average and logarithmic transformation  mainly for the reason that it allows us to drive an analytical expression for our loss function which can be directly and efficiently optimized by our deep neural network during the learning process. Another motivation for the logarithmic function is that it converts (\ref{eq-org-loss}) into the form of maximum log-likelihood estimation, which is extensively used in machine learning and pattern recognition analysis. In Section \ref{sec-averaging}, we will also evaluate other options of averaging operations. Our experimental results will show that the proposed solution achieves the best performance. 

Combining (\ref{eq-ha}) with (\ref{eq-org-loss}), we have the following minimization problem:
 \begin{equation}
 \begin{split}
     \min \mathcal{L}_1 &= \mathbb{E}^A_{\bm{x}_i\in\mathcal{X}}\log\{\frac{1}{|\mathcal{P}_i|}\sum_{\bm{x_j}\in \mathcal{P}_i} \frac{1}{p[l_{\bm{x}_i} \approx l_{\bm{x}_j}\ |\ m(\bm{x}_i, \bm{x}_j)]}\}\\
     &+\mathbb{E}^A_{\bm{x}_i\in\mathcal{X}}\log\{\frac{1}{|\mathcal{N}_i|}\sum_{\bm{x_j}\in \mathcal{N}_i} \frac{1}{p[l_{\bm{x}_i} \not\approx l_{\bm{x}_j}\ |\ m(\bm{x}_i, \bm{x}_j)]}\}.
\end{split}
\label{eq-org-loss-1}
\end{equation}
It should be noted that these two  probabilities for positive and negative pairs conditioned by the metric function are both posterior probabilities, which cannot be directly obtained from the network inference during metric learning. In other words, this loss function cannot be directly optimized during deep metric learning. To address this issue, in the following section, we propose to transform these posterior probabilities into \textit{a prior} probabilities using the following contrastive Bayesian analysis. 

\subsection{Contrastive Bayesian Analysis}
\label{se:bayesian-analysis}
In this section, we will perform constrastive Bayesian analysis of the optimization problem formulated in (\ref{eq-org-loss-1}) so that it can be use to guide the optimization of a deep neural network. 
According to the Bayesian property, we have
\begin{equation}
\begin{split}
    &p[l_{\bm{x}_i} \approx l_{\bm{x}_j}\ |\ m(\bm{x}_i, \bm{x}_j)] \\
    = &\frac{p[m(\bm{x}_i, \bm{x}_j)\ |\ l_{\bm{x}_i} \approx l_{\bm{x}_j}]\cdot p[l_{\bm{x}_i} \approx l_{\bm{x}_j}]}{p[m(\bm{x}_i, \bm{x}_j)]},
    \end{split}
    \label{eq:bayes-p}
\end{equation}
and
\begin{equation}
\begin{split}
      &p[l_{\bm{x}_i} \not\approx l_{\bm{x}_j}\ |\ m(\bm{x}_i, \bm{x}_j)] \\
    = &\frac{p[m(\bm{x}_i, \bm{x}_j)\ |\ l_{\bm{x}_i} \not\approx l_{\bm{x}_j}]\cdot p[l_{\bm{x}_i} \not\approx l_{\bm{x}_j}]}{p[m(\bm{x}_i, \bm{x}_j)]}.
\end{split}
    \label{eq:bayes-n}
\end{equation}
Dividing  (\ref{eq:bayes-p}) by (\ref{eq:bayes-n}), we have
\begin{equation}
\begin{aligned}
    &\frac{p[l_{\bm{x}_i} \approx l_{\bm{x}_j}\ |\ m(\bm{x}_i, \bm{x}_j)]}{p[l_{\bm{x}_i} \not\approx l_{\bm{x}_j}\ |\ m(\bm{x}_i, \bm{x}_j)]}\\
    =&\frac{p[m(\bm{x}_i, \bm{x}_j)\ |\ l_{\bm{x}_i} \approx l_{\bm{x}_j}]}{p[m(\bm{x}_i, \bm{x}_j)\ |\ l_{\bm{x}_i} \not\approx l_{\bm{x}_j}]}\cdot\frac{p[l_{\bm{x}_i} \approx l_{\bm{x}_j}]}{p[l_{\bm{x}_i} \not\approx l_{\bm{x}_j}]}.
\end{aligned}
\label{eq:bayes-divide}
\end{equation}
Define
\begin{equation}
    \Phi_N[m(\bm{x}_i,\bm{x}_j)] = \frac{p[m(\bm{x}_i, \bm{x}_j)\ |\ l_{\bm{x}_i} \approx l_{\bm{x}_j}]}{p[m(\bm{x}_i, \bm{x}_j)\ |\ l_{\bm{x}_i}\not\approx l_{\bm{x}_j}]}
\end{equation}
and
\begin{equation}
    \Phi_P[m(\bm{x}_i,\bm{x}_j)] = \frac{p[m(\bm{x}_i, \bm{x}_j)\ |\ l_{\bm{x}_i} \not\approx l_{\bm{x}_j}]}{p[m(\bm{x}_i, \bm{x}_j)\ |\ l_{\bm{x}_i} \approx l_{\bm{x}_j}]}.
\end{equation}
Note that the ratio between the probability of a pair of samples to be in the positive set and the probability of the pair in the negative set is equal to the size ratio $\bm{\Theta}_{PN}$ between these two sets. In other words,
\begin{equation}
    \frac{p[l_{\bm{x}_i} \approx l_{\bm{x}_j}]}{p[l_{\bm{x}_i} \not\approx l_{\bm{x}_j}]}=\frac{|\mathcal{P}_i|}{|\mathcal{N}_i|} \overset{def}{=} \bm{\Theta}_{PN}.
\end{equation}
According to (\ref{eq:pos-neg}) and (\ref{eq:bayes-divide}), we have 
\begin{equation}
\begin{split}
    &\frac{1-p[l_{\bm{x}_i} \not\approx l_{\bm{x}_j}\ |\ m(\bm{x}_i, \bm{x}_j)]}{p[l_{\bm{x}_i} \not\approx l_{\bm{x}_j}\ |\ m(\bm{x}_i, \bm{x}_j)]}
    =\Phi_N[m(\bm{x}_i,\bm{x}_j)] \cdot \bm{\Theta}_{PN},
\end{split}
    \nonumber
\end{equation}
which yields
\begin{equation}
p[l_{\bm{x}_i} \not\approx l_{\bm{x}_j}\ |\ m(\bm{x}_i, \bm{x}_j)]
    =\frac{1}{1+\Phi_N[m(\bm{x}_i,\bm{x}_j)]\cdot\bm{\Theta}_{PN}}.
    \label{eq:posterior-n}
\end{equation}
Similarly, we have
\begin{equation}
    p[l_{\bm{x}_i} \approx l_{\bm{x}_j}\ |\ m(\bm{x}_i, \bm{x}_j)]
    =\frac{1}{1+\Phi_P[m(\bm{x}_i,\bm{x}_j)]\cdot\bm{\Theta}_{PN}^{-1}}.
    \label{eq:posterior-p}
\end{equation}
Inserting (\ref{eq:posterior-n}) and (\ref{eq:posterior-p}) into the objective function in our optimization problem (\ref{eq-org-loss-1}), we have 
\begin{equation}
\begin{split}
  \min \mathcal{L}_1 & = \mathbb{E}_{\bm{x}_i\in\mathcal{X}}^A\log\left\{1 + \frac{1}{|\mathcal{P}_i|}\sum_{\bm{x}_j\in \mathcal{P}_i} \Phi_P[m(\bm{x}_i, \bm{x}_j)]\cdot \bm{\Theta}_{PN}^{-1}\right\} \\
  &+ \mathbb{E}_{\bm{x}_i\in\mathcal{X}}^A\log\left\{1 + \frac{1}{|\mathcal{N}_i|} \sum_{\bm{x}_j\in \mathcal{N}_i} \Phi_N[m(\bm{x}_i, \bm{x}_j)]\cdot \bm{\Theta}_{PN}\right\}.
\end{split}
\label{eq:single-bayes}
\end{equation}

\begin{figure*}[t]
	\begin{center}
		\includegraphics[width=0.8\linewidth]{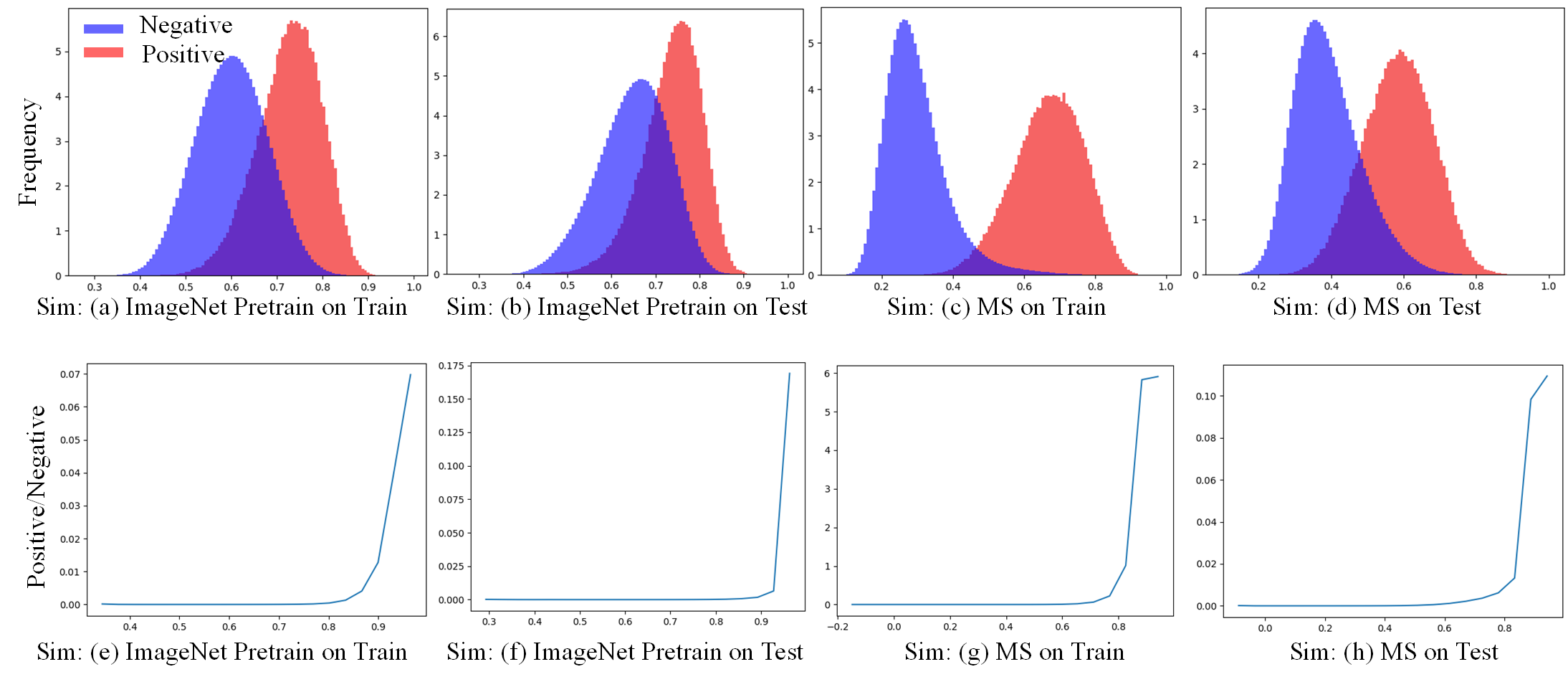}
	\end{center}
	\vspace{-4mm}
	\caption{The similarity distributions on the CUB training and test sets with the ImageNet pretrained model and the model trained by the MS method \cite{CVPR-WangHHDS19} based on the GoogLeNet backbone network. (a) and (b) are the similarity distributions based on features extracted from the ImageNet pretrained model, (e) and (f) are the corresponding ratio functions of positive and negative sample distributions. (c) and (d) are the similarity distributions based on features learned with the MS method, (g) and (h) are the corresponding ratio functions of positive and negative sample distributions.}
	\label{fig:CUB}
	\vspace{-4mm}
\end{figure*}

Next, we develop analytical models for 
$\Phi_N[m(\bm{x}_i, \bm{x}_j)] $ and 
$\Phi_P[m(\bm{x}_i, \bm{x}_j)]$, which are ratios between the probabilities of negative and positive pairs conditioned by the metric function $ m(\bm{x}_i, \bm{x}_j)$. We observe that both of these two conditional probabilities 
$p[m(\bm{x}_i, \bm{x}_j)\ |\ l_{\bm{x}_i} \approx l_{\bm{x}_j}]$ and 
$p[m(\bm{x}_i, \bm{x}_j)\ |\ l_{\bm{x}_i} \not\approx l_{\bm{x}_j}]$ follow approximate Gaussian distributions, which is a very natural assumption for the distribution of the metric function $ m(\bm{x}_i, \bm{x}_j)$. 
For example, Fig. \ref{fig:CUB}
shows the conditional probability distribution of positive pairs $p[m(\bm{x}_i, \bm{x}_j)\ |\ l_{\bm{x}_i} \approx l_{\bm{x}_j}]$ and the conditional probability distribution of negative pairs $p[m(\bm{x}_i, \bm{x}_j)\ |\ l_{\bm{x}_i} \not\approx l_{\bm{x}_j}]$. We can see that they are approximately Gaussian distributions. 
Specifically, 
\begin{equation}
\begin{split}
   p[m(\bm{x}_i, \bm{x}_j) \ |\ l_{\bm{x}_i} &\approx l_{\bm{x}_j}]= \frac{1}{\sigma_P\sqrt{2\pi}} e^{-(z -\mu_P)^2/(2\sigma_P^2)} \\
   p[m(\bm{x}_i, \bm{x}_j)\ |\ l_{\bm{x}_i} &\not\approx l_{\bm{x}_j} ]=\frac{1}{\sigma_N\sqrt{2\pi}} e^{-(z -\mu_N)^2/(2\sigma_N^2)},    
\end{split}
\nonumber
\end{equation}
where $z=m(\bm{x}_i, \bm{x}_j)$. It should be pointed out that this phenomenon of Gaussian distribution is not metric-dependent, it is general for all cases. Because more pairwise distances or similarities between features of image samples are always concentrated near the metric mean. Then, we have
\begin{equation}
\begin{split}
\Phi_N[m(\bm{x}_i,\bm{x}_j)] &= \frac{p[m(\bm{x}_i, \bm{x}_j)\ |\ l_{\bm{x}_i} \approx l_{\bm{x}_j}]}{p[m(\bm{x}_i, \bm{x}_j)\ |\ l_{\bm{x}_i}\not\approx l_{\bm{x}_j}]} \\
&=\frac{\frac{1}{\sigma_P}e^{-(z -\mu_P)^2/(2\sigma_P^2)} }{
\frac{1}{\sigma_N}e^{-(z -\mu_N)^2/(2\sigma_N^2)} }\\
&=\frac{\sigma_N}{\sigma_P}e^{(z -\mu_N)^2/(2\sigma_N^2)-(z -\mu_P)^2/(2\sigma_P^2)}\\
&=\frac{\sigma_N}{\sigma_P}e^{\frac{z^2-2\mu_N\cdot z +\mu_N^2}{2\sigma_N^2}-\frac{z^2-2\mu_P\cdot z +\mu_P^2}{2\sigma_P^2}}\\
&=\varsigma \cdot e^{\zeta_1\cdot z^2+\zeta_2 \cdot z+\zeta_3}.
\end{split}
\end{equation}
where $\varsigma=\frac{\sigma_N}{\sigma_P}$, $\zeta_1=\frac{1}{2\sigma_N^2}-\frac{1}{2\sigma_P^2}$, $\zeta_2=\frac{\mu_P}{\sigma_P^2}-\frac{\mu_N}{\sigma_N^2}$, $\zeta_3=\frac{\mu_N^2}{2\sigma_N^2}-\frac{\mu_P^2}{2\sigma_P^2}$ are constants. We have two different cases. In the first case, $\sigma_P^2 = \sigma_N^2 = \sigma^2$, we have
\begin{equation}
\begin{split}
\Phi_N[m(\bm{x}_i,\bm{x}_j)]
&= e^{[2(\mu_P - \mu_N)\cdot z + (\mu_N^2 -\mu_P^2)] / (2\sigma^2) }.
\end{split}
\label{eq:linear}
\end{equation}
This indicates that $\Phi_N[m(\bm{x}_i,\bm{x}_j)]$ follows an exponential function. In the second case, $\sigma_P^2 \not\approx \sigma_N^2$, we have
\begin{equation}
\begin{split}
\Phi_N[m(\bm{x}_i,\bm{x}_j)]
&= \varsigma \cdot e^{\zeta_1\cdot z^2+\zeta_2 \cdot z+\zeta_3}.
\end{split}
\label{eq:quadratic}
\end{equation}
In this case, the exponent is a quadratic function. We plot the ratio function of positive and negative sample distributions $\Phi_N[m(\bm{x}_i,\bm{x}_j)]$ in Fig. \ref{fig:CUB}. We can see that $\Phi_N[m(\bm{x}_i,\bm{x}_j)]$ can be approximated by a general exponential function:
\begin{equation}
\begin{split}
\Phi_N[m(\bm{x}_i,\bm{x}_j)]
&= e^{\zeta \cdot z-\zeta_0}.
\end{split}
\label{eq:linear-simple}
\end{equation}
To further verify the effectiveness of this approximation, we  train models on the CUB dataset with linear and quadratic exponents using the ResNet-50 backbone, respectively. Results showed that their performance are similar, ranging between 69.5\% and 70.0\%.

According to the above analysis, for the convenience of derivation, $\Phi_N[m(\bm{x}_i,\bm{x}_j)]$ can be approximated by the following  exponential function:
\begin{equation}
    \Phi_N[m(\bm{x}_i, \bm{x}_j)] = \exp\left\{\frac{m(\bm{x}_i,\bm{x}_j)-\alpha^N}{\beta^N}\right\}.
    \nonumber
\end{equation}
Similarly, we have 
\begin{equation}
    \Phi_P[m(\bm{x}_i, \bm{x}_j)] = \exp\left\{\frac{\alpha^P-m(\bm{x}_i,\bm{x}_j)}{\beta^P}\right\}.
    \nonumber
\end{equation}
$\alpha^N$, $\beta^N$, $\alpha^P$, and $\beta^P$ are parameters for the exponential functions.
With these, (\ref{eq:single-bayes}) can be rewritten as follows
\begin{equation}
\begin{split}
  \min \mathcal{L}_1 & = \mathbb{E}_{\bm{x}_i\in\mathcal{X}}\log\left\{1 + \delta^P \sum_{\bm{x}_j\in \mathcal{P}_i} \exp[\frac{\alpha^P-m(\bm{x}_i,\bm{x}_j)}{\beta^P}]\right\} \\
  &+ \mathbb{E}_{\bm{x}_i\in\mathcal{X}}\log\left\{1 + \delta^N \sum_{\bm{x}_j\in \mathcal{N}_i} \exp[\frac{m(\bm{x}_i,\bm{x}_j)-\alpha^N}{\beta^N}]\right\},
\end{split}
\label{eq:log-bayes-transform}
\end{equation}
where 
\begin{equation}
    \delta^P = \frac{|\mathcal{N}_i|}{|\mathcal{P}_i|^2}, \quad
    \delta^N = \frac{|\mathcal{P}_i|}{|\mathcal{N}_i|^2}.
\end{equation}

\subsection{Other Choices of Averaging Operations and Loss Functions}
\label{sec-averaging}
 
 In (\ref{eq-org-loss-1}), we choose the
 harmonic average followed by log-average to formulate the average conditional probabilities. 
 Certainly, there are many other choices to perform this averaging operations. 
 A generic averaging operation can be defined as follows
 \begin{equation}
     \mathbb{E}[\bm{x}] = f^{-1}\left[\frac{1}{|\mathcal{X}|}
     \sum_{\bm{x}_i\in \mathcal{X}} f(\bm{x}_i)\right].
     \label{eq-avg-options}
 \end{equation}
 For  the  loss function in (\ref{eq-org-loss-1}), 
 $f(x) = \log(x)$.
 In our experiments, we have also evaluated two other choices of $f(x)$. The first one is $f(x)  = x$ whose corresponding loss function is denoted by $\mathcal{L}_{const}$.
 The second choice is the square root function $f(x)=\sqrt{x}$ whose loss function is denoted by $\mathcal{L}_{sqrt}$.
 Our experimental results will demonstrate that the original choice of $f(x) = \log(x)$ yields the best performance in deep metric learning.
 
\subsection{Learning with Hard Positive and Negative Pairs}
According to recent studies on deep metric learning \cite{CVPR-WangHHDS19}, it is beneficial to use the statistics of hard samples, specifically, hard positive and negative pairs, instead of all positive and negative pairs, to define the loss function for network training. 
In our formulation, positive pairs should have large similarity values while negative pairs should have small similarity values.
This implies that the similarity values of typical positive pairs should be larger than those of the typical negative pairs. If a positive pair has a similarity value even smaller than the typical similarity of negative pairs, then this is a hard positive pairs which the network learning should pay extra attention to. 
Similarly, if a negative pair has a similarity value even larger than those of the positive pairs, then this is a hard negative pair. 
Specifically, given a mini-batch, for sample $\bm{x}_i$, we define its set of hard positive pairs  $\mathcal{P}_i^*$
and set of hard negative pairs $\mathcal{N}_i^*$
as 
\begin{equation}
\begin{split}
    &\mathcal{P}_i^*=\{\bm{x}_j\in \mathcal{P}_i\ |\ m(\bm{x}_i, \bm{x}_j)< \Gamma^{max}_{\mathcal{N}_i} +\epsilon\} \\
    &\mathcal{N}_i^*=\{\bm{x}_j\in \mathcal{N}_i\ |\ m(\bm{x}_i, \bm{x}_j)> \Gamma^{min}_{\mathcal{P}_i}-\epsilon\}
    \label{eq:hard-negative-positive-pairs}
\end{split}
\end{equation}
where 
\begin{equation}
\begin{split}
    \Gamma^{max}_{\mathcal{N}_i} & = max\{m(\bm{x}_i, \bm{x}_j)\ |\ \bm{x}_j\in \mathcal{N}_i\}, \\
    \Gamma^{min}_{\mathcal{P}_i} & = min\{m(\bm{x}_i, \bm{x}_j)\ |\ \bm{x}_j\in \mathcal{P}_i\},
\end{split}
\end{equation}
and $\epsilon$ is a marginal threshold to control the number of selected examples. Then, the loss function of (\ref{eq:log-bayes-transform}) based on hard positive pairs and hard negative pairs is computed as follows:
\begin{equation}
\begin{split}
\mathcal{L}_1=&\mathbb{E}_{\bm{x}_i\in\mathcal{X}}\log\{1+\delta^P\sum_{\bm{x}_j \in \mathcal{P}_i^*}\exp[\frac{\alpha^{P}-m(\bm{x}_i, \bm{x}_j)}{\beta^{P}}]\}\\
    +&\mathbb{E}_{\bm{x}_i\in\mathcal{X}}\log\{1+\delta^N\sum_{\bm{x}_j \in \mathcal{N}_i^*}\exp[\frac{m(\bm{x}_i, \bm{x}_j)-\alpha^{N}}{\beta^{N}}]\}.
   \end{split}
    \label{eq:CBML-hard}
\end{equation}
Note that the optimization of deep learning algorithm is usually based on examples of mini-batches. Thus, the hard positive and negative pairs are selected from the current mini-batch.

\subsection{Deep Metric Learning with Generalization Constraints}

In the above section, we have successfully established a new loss function for deep metric learning which bridges the semantic gap between the metric function and class labels using contrastive Bayesian analysis. 
According to our experiments, this new metric learning method performs very well on the training set, but it may not generalize well onto the test set which consists of totally new classes. To improve its generalization capability, we propose to derive a generalization constraint and incorporate it into the contrastive Bayesian analysis.

\begin{figure}[t]
	\begin{center}
		\includegraphics[width=0.9\linewidth]{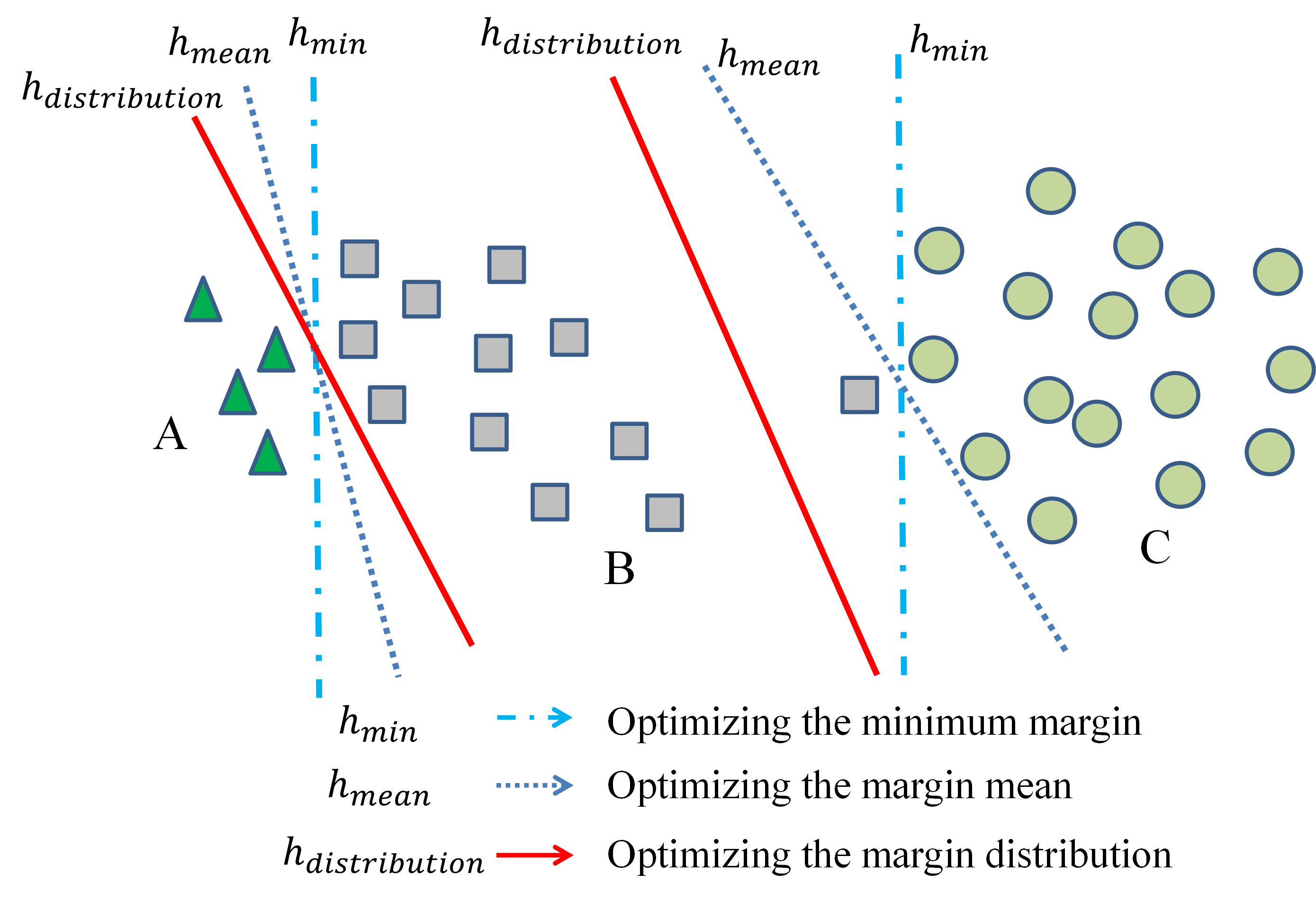}
	\end{center}
	\vspace{-4mm}
	\caption{Illustration of linear separators with outliers or noisy data points.}
	\label{fig:margin}
	\vspace{-4mm}
\end{figure}

Successful generalization of the network model learned from the training set to the test set is an important problem in machine learning \cite{ANNPR-Zhou14, AI-GaoZ13a}. In traditional classification with SVM (support vector machine) \cite{ML-CortesV95} and AdaBoost \cite{ECCCLT-FreundS95} classifiers, Zhou \textit{et al.} \cite{ANNPR-Zhou14, AI-GaoZ13a, KDD-ZhangZ14, TKDE-ZhangZ20} have pointed out that optimizing the marginal distribution by maximizing the marginal mean and minimizing the marginal variance simultaneously can lead to better generalization performance. As shown in Fig.\ref{fig:margin}, samples in class A (denoted by triangles) and B (squares), and examples in class B and C (circles) are linearly separable by hyperplanes of $h_{\min}$, $h_{\text{mean}}$ and $h_{\text{distribution}}$. The hyperplane $h_{\min}$ is obtained by minimizing the smallest similarity or margin between samples of these two classes, $h_{\text{mean}}$ is obtained by minimizing the mean margin between samples from these two classes. 
These two classifiers may over-fit the training data and are sensitive to outliers. For example, the outlier sample in class B will move the hyperplanes  $h_{\min}$ and $h_{\text{mean}}$ very close to class C, resulting in degraded generalization capability.
To address this issue, the $h_{\text{distribution}}$ classifier considers all samples instead of the extreme samples  to avoid being affected by individual outliers. It aims to  optimize the margin distribution by maximizing the margin mean and minimize the margin variance. Here, the margin represents the distance between the sample and the classification hyperplane \cite{ANNPR-Zhou14}.

\begin{figure}[t]
	\begin{center}
		\includegraphics[width=0.9\linewidth]{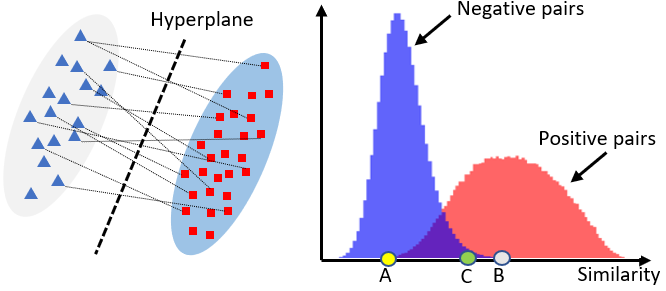}
	\end{center}
	\vspace{-4mm}
	\caption{Illustrate the main idea of our metric variance constraint.}
	\vspace{-4mm}
	\label{fig:constraint-distribution}
\end{figure}

Motivated by this observation, in this work, we propose a metric variance constraint based on similarities of features into our contrastive Bayesian analysis for deep metric learning to improve the model generalization capability. Comparing to the traditional classifier design where the features are pre-computed by the feature extraction method, the unique challenge in deep metric learning is that the feature here is also learned on the fly in an end-to-end manner.
In our contrastive Bayesian analysis, we use the contrastive Bayesian loss between positive and negative pairs to define a loss function to train the network model. We do not have the explicit hyperplane in the optimization process. 
How do we build the metric variance constraint into  this pairwise contrastive Bayesian loss? 
Our main idea is illustrated in  Fig. \ref{fig:constraint-distribution}.
We recognize that, when we minimize the marginal variance of all samples with respect to the classification hyperplane, the similarity metric 
$m(\bm{x}_i, \bm{x}_j)$ between negative pairs of two classes should aggregate towards a similarity hyperplane. For example, if negative pairs of two classes have very similar $m(\bm{x}_i, \bm{x}_j)$, then samples from these two classes should have similar distance to the hyperplane. Based on this observation, in our proposed method, we first  calculate a target value corresponding to $\bm{x}_i$ and then push the feature similarity 
$m(\bm{x}_i, \bm{x}_j)$ between negative pairs of two classes towards this target value so as to minimize their metric variance. These  target values form the similarity hyperplane.
From our experiments, we find out that points near the average value of $m(\bm{x}_i, \bm{x}_j)$ of all negative pairs between two classes is a good choice for the target value. 

\begin{figure*}[t]
	\begin{center}
		\includegraphics[width=0.8\linewidth]{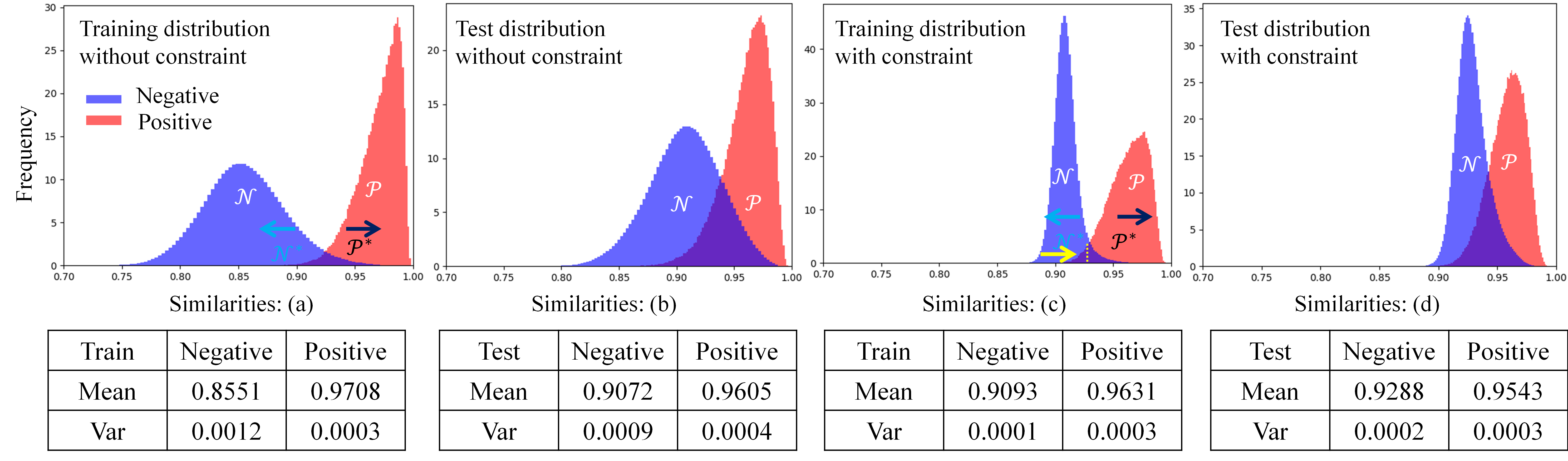}
	\end{center}
	\vspace{-4mm}
	\caption{The similarity distributions on the training and test sets with and without metric variance constraint (MVC) based on the ResNet-50 backbone network on the CUB dataset. (a) and (b) are the similarity distributions on the training and test set without using the MVC. (c) and (d) are similarity distributions on the training and test set with the MVC.}
	\label{fig:constraint}
\end{figure*}

Specifically, we define the target value $\xi_i$ in two classes for sample $\bm{x}_i$ as
\begin{equation}
    \xi_i = \gamma \cdot \mathbb{E}_{\mathcal{P}_i}
    \{m(\bm{x}_i,\bm{x}_j)\} + (1-\gamma) \cdot \mathbb{E}_{\mathcal{N}_i} 
    \{m(\bm{x}_i,\bm{x}_j)\},
    \label{eq:hyperplane}
\end{equation}
which is the weighted average $m(\bm{x}_i,\bm{x}_j)$ of negative pairs (example point A  in Fig. \ref{fig:constraint-distribution}) and positive pairs (example point B in Fig. \ref{fig:constraint-distribution}), respectively. 
The selection of the control parameter $\gamma$ will be evaluated in our ablation studies. During training, $\xi_i$ is updated for each sample $\bm{x}_i$.

With the definition of the target value, we 
introduce the following metric variance loss
\begin{equation}
\begin{split}
    \mathcal{L}_2&=\mathbb{E}_{\bm{x}_i\in\mathcal{X}}\mathbb{E}_{\bm{x}_j\in\mathcal{N}_i}[m(\bm{x}_i,\bm{x}_j)-\xi_i]^2\\
    &=\mathbb{E}_{\bm{x}_i\in\mathcal{X}}\left\{\frac{1}{|\mathcal{N}_i|}\sum_{\bm{x}_j\in\mathcal{N}_i}[m(\bm{x}_i,\bm{x}_j)-\xi_i]^2\right\},
\end{split}
    \label{eq:constraint}
\end{equation}
which minimizes the variance of $m(\bm{x}_i, \bm{x}_j)$ for all negative pairs of any two classes, aiming to improve the generalization capability of our learned model. During training, the negative pairs share the same $\bm{x}_i$ in a mini-batch.
Incorporating this metric variance loss into the contrastive Bayesian loss in (\ref{eq:CBML-hard}), we have the following loss function
\begin{equation}
\begin{split}
\quad \mathcal{L}=&\mathcal{L}_1+\lambda\mathcal{L}_2\\
=&\mathbb{E}_{\bm{x}_i\in\mathcal{X}}\log\{1+\delta^P\sum_{\bm{x}_j \in \mathcal{P}^*_i}\exp[\frac{\alpha^{P}-m(\bm{x}_i, \bm{x}_j)}{\beta^{P}}]\}\\
    +&\mathbb{E}_{\bm{x}_i\in\mathcal{X}}\log\{1+\delta^N\sum_{\bm{x}_j \in \mathcal{N}^*_i}\exp[\frac{m(\bm{x}_i, \bm{x}_j)-\alpha^{N}}{\beta^{N}}]\}\\
   +&\lambda \cdot \mathbb{E}_{\bm{x}_i\in\mathcal{X}}\frac{1}{|\mathcal{N}_i|}\sum_{\bm{x}_j\in\mathcal{N}_i}[m(\bm{x}_i,\bm{x}_j)-\xi_i]^2.
   \end{split}
    \label{eq:CBML}
\end{equation}

To understand the impact of the metric variance loss, we plot the distributions of $m(\bm{x}_i, \bm{x}_j)$ for positive and negative pairs on the training and test sets of the CUB dataset  in Fig. \ref{fig:constraint}.
(a) shows the distributions of $m(\bm{x}_i, \bm{x}_j)$ of the training samples without using the metric variance constraint. 
(b) shows the distributions of the test samples. 
(c) and (d) shows the distributions with the metric variance constraint. 
We can see that the metric variance loss makes the similarity distribution of negative pairs much more compact, effectively reducing the metric variance of all samples and improving the generalization capability of our deep metric learning method.

\subsection{Summary of Algorithm}
The proposed contrastive Bayesian metric learning (CBML) method is summarized in Algorithm \ref{algorithm:overview}.

\begin{algorithm}
\caption{Summary of Optimization Algorithm}
\label{algorithm:overview}
\begin{algorithmic}[1]

\State \textbf{Initialization:} Initialize network parameters, the value of margin threshold $\epsilon$, learning rate, and mini-batch size $o$.
    
\State \textbf{Input:}  Training images.
  
\State \textbf{Output:} Optimized model.
      \State \textbf{Step 1:} Randomly sample a mini-batch images with corresponding labels from the training dataset.
      \State \textbf{Step 2:} Calculate features of the sampled mini-batch images by the backbone encoder.
      \State \textbf{Step 3:} Calculate loss based on (\ref{eq:CBML}).
      \State \hspace{1.0cm} \textbf{for} $i$ in range($o$) \textbf{do}
        \State \hspace{1.5cm} Make up sets of $\mathcal{P}_i^*$, and $\mathcal{N}_i^*$ based on their\\ \hspace{1.5cm} definitions in (\ref{eq:hard-negative-positive-pairs}).
        \State \hspace{1.5cm} Calculate contrastive Bayesian loss based on (\ref{eq:CBML-hard}).
        \State \hspace{1.5cm} Calculate similarity hyperplane and marginal sim-\\ \hspace{1.5cm} ilarity variance based on (\ref{eq:hyperplane}) and (\ref{eq:constraint}).
     \State \hspace{1.0cm} \textbf{end for}
     \State \textbf{Step 4:} Back propagate and update parameters.
  \end{algorithmic}
\end{algorithm}

\section{Experimental Results}

In the following experiments, following the same procedure used by existing papers \cite{CVPR-SongXJS16,PAMI-Opitz18,TIP-KanCHZZW19, PAMI-Milbich20}, we evaluate the performance of the proposed CBML method.

\subsection{Datasets}
The following benchmark datasets are used in our experiments. For all the experiments on these datasets, the training classes and test classes are different.
(1) The \textbf{CUB-200-2011} \cite{CIT-wah2011} consists of 11,788 images from 200 bird categories. We use the first 100 classes (5,864 images) for training and the remaining 100 classes (5,924 images) for testing.
(2) The \textbf{Cars-196} \cite{ICCVW-Krause0DF13} dataset contains 16,185 images of 196 cars classes. We use the first 98 classes (8,054 images) for training and the remaining 98 classes (8,131 images) for testing.
(3) The \textbf{Stanford Online Product (SOP)} \cite{CVPR-SongXJS16} dataset consists of 120,053 images with 22,634 classes crawled from Ebay. Classes are hierarchically grouped into 12 coarse categories (e.g. cup, bicycle, etc.). Following the existing protocol, we split the first 11,318 classes with 59,551 images for training, and the remaining 11,316 classes with 60,502 images for retrieval. In the test set, each image is also used as the query image.
(4) The \textbf{In-Shop Clothes Retrieval (In-Shop)} \cite{CVPR-LiuLQWT16} dataset consists of 52,712 images with 7,986 clothing classes. We use the predefined 25,882 training images of 3,997 classes for training. The remaining 3985 classes are partitioned into a query set (14,218 images) and a gallery set (12,612 images).
(5) The \textbf{ImageNet ILSVRC 2012} \cite{IJCV-RussakovskyDSKS15} dataset contains 1,000 classes with 1,281,167 training images and 50,000 validation images. In order to verify the generalization ability of our algorithm on general images, we split the first 500 classes with 667,289 images in training and validation sets as our training set, and the remaining 500 classes with 638,878 images in training set and with 25,000 images in validation set are used for our gallery set and query set, respectively.

\begin{table*}[h]
		\caption{The comparison results on the CUB-200-2011 \cite{CIT-wah2011} and the Cars-196 \cite{ICCVW-Krause0DF13} datasets. It should be noted that not all methods provided results for all the evaluation metrics. In this case, they are left blank in the tables. }
		\begin{center}
		\resizebox{\linewidth}{!}{
			\begin{tabularx}{17.8cm}{c|l|c|ccccc||ccccc}
				\hline 
				\multirow{2}[1]*{Backbone} & \multirow{2}[1]*{Methods}&\multirow{2}[1]*{Dim}&\multicolumn{5}{c||}{\textbf{CUB}} &\multicolumn{5}{c}{\textbf{Cars}}\\
				\cline{4-13} 
				& & & R@1 & R@2 & R@4 & R@8 &NMI& R@1 & R@2 & R@4 & R@8 &NMI \\
				\hline
				\hline
				\multirow{14}[5]*{GoogLeNet} & Contrastive [CVPR16] \cite{CVPR-SongXJS16} & 128 & 26.4 & 37.7 & 49.8 & 62.3 &46.1 & 21.7 &32.3 & 46.1& 58.9&48.0\\
				&Triplet [CVPR16] \cite{CVPR-SongXJS16}& 128 & 36.1& 48.6& 59.3& 70.0 &49.8& 39.1& 50.4& 63.3& 74.5&52.9\\
				& HDC [ICCV17] \cite{ICCV-YuanYZ17} & 384 & 53.6 &65.7& 77.0& 85.6&- & 73.7& 83.2& 89.5& 93.8&-\\
				&LiftedStruct [CVPR16] \cite{CVPR-SongXJS16} & 512 & 47.2& 58.9& 70.2& 80.2& 56.5 & 49.0& 60.3& 72.1& 81.5& 56.9\\
				&Binomial Deviance [NIPS16] \cite{NIPS-UstinovaL16} & 512& 52.8& 64.4& 74.7& 83.9&- & - &- & - &-&-\\
				& Histogram Loss [NIPS16] \cite{NIPS-UstinovaL16} & 512 & 50.3& 61.9& 72.6& 82.4& -&- &- &- & -&-\\
				& N-Pair-Loss [NIPS16] \cite{NIPS-Sohn16} & 512 &51.0& 63.3& 74.3& 83.2& 60.4 &71.1&79.7&86.5&91.6&64.0\\
				& Angular Loss [ICCV17] \cite{ICCV-WangZWLL17} & 512& 54.7& 66.3& 76.0& 83.9& 61.1& 71.4& 81.4& 87.5& 92.1&63.2\\
				& BIER [ICCV17] \cite{ICCV-OpitzWPB17} & 512 &55.3& 67.2& 76.9& 85.1&- & 78.0& 85.8& 91.1& 95.1&-\\
				& HDML [\textcolor{blue}{CVPR19}] \cite{CVPR-ZhengCL019}& 512 & 53.7 & 65.7 & 76.7 & 85.7& 62.6& 79.1& 87.1& 92.1 & 92.5&\textbf{69.7}\\
				&MS [\textcolor{blue}{CVPR19}] \cite{CVPR-WangHHDS19}& 512 & 57.4  & 69.8 & 80.0 & 87.8&-& 77.3 & 85.3 & 90.5 & 94.2&-\\
				& A-BIER [\textcolor{blue}{TPAMI20}] \cite{PAMI-Opitz18} & 512& 57.5  & 68.7 & 78.3 & 82.6&- & 82.0 & 89.0 & 93.2 & 96.1&- \\
				&MS+EE [\textcolor{blue}{CVPR20}] \cite{cvpr20-KoG20}& 512& 57.4 & 68.7 & 79.5 & 86.9& 63.3& 76.1 & 84.2 & 89.8 & 93.8&63.5\\
				&MS+XBM [\textcolor{blue}{CVPR20}] \cite{CVPR-Wang20}& 512 & 61.9 & 72.9 & 81.2 & 88.6&- & 80.3 & 87.1 & 91.9 & 95.1&-\\
				&LoOp+MS [\textcolor{blue}{ICCV21}] \cite{Vasudeva-2021-ICCV} & 512 & 52.0 & 64.3 & 75.0 & 84.1 & 61.1& 72.6 &81.5& 88.4 & 92.8 & 63.0\\
				\cline{2-13}
				& \textbf{Ours}: CBML & 512 & \textbf{63.8} & \textbf{74.8} & \textbf{83.6} & \textbf{90.3}& \textbf{66.7} & \textbf{83.0} & \textbf{89.3} & \textbf{93.5} & \textbf{96.2}& 68.8\\
				\hline
				\multirow{11}[4]*{BN-Inception}
				&RLL-H [\textcolor{blue}{CVPR19}] \cite{CVPR-WangHKHGR19} & 512& 57.4 & 69.7 & 79.2 & 86.9& 63.6& 74.0 & 83.6 & 90.1 & 94.1&65.4\\
				&RLL-(L,M,H) [\textcolor{blue}{CVPR19}] \cite{CVPR-WangHKHGR19} &1536 &61.3 & 72.7 & 82.7 & 89.4& 66.1 & 82.1 & 89.3 & 93.7 & 96.7&71.8\\
				&SoftTriple [\textcolor{blue}{ICCV19}] \cite{ICCV-Qian2019} &512 & 65.4 & 76.4 & 84.5 & 90.4& 69.3 & 84.5 & 90.7 & 94.5 & 96.9& 70.1\\
				&DeML [\textcolor{blue}{CVPR19}] \cite{CVPR-ChenD19} & 512&  65.4 & 75.3 & 83.7 & 89.5&- & 86.3 & 91.2 & 94.3 & 97.0&-\\
				&MS [\textcolor{blue}{CVPR19}] \cite{CVPR-WangHHDS19} &512 &65.7 & 77.0 & 86.3 & 91.2& -& 84.1 & 90.4 & 94.0 & 96.5&-\\
				&Contrastive+HORDE [\textcolor{blue}{ICCV19}] \cite{ICCV-Jacob2019} &512 &66.8 & 77.4 & 85.1 & 91.0& -& 86.2 & 91.9 & 95.1 & 97.2&-\\
				&MS+XBM [\textcolor{blue}{CVPR20}] \cite{CVPR-Wang20}& 512 &65.8 & 75.9 & 84.0 & 89.9&-& 82.0 & 88.7 & 93.1 & 96.1&-\\
				& Proxy-Anchor [\textcolor{blue}{CVPR20}] \cite{CVPR20-KimKCK20} & 512 & 68.4 & 79.2 & 86.8 & 91.6&- &  86.1 & 91.7& 95.0 & 97.3&- \\
				&DRML-PA [\textcolor{blue}{\footnotesize{ICCV21}}] \cite{ICCV-Zheng21}& 512 & 68.7 & 78.6 & 86.3 & 91.6& 69.6 &  86.9 & 92.1& \textbf{95.2} & \textbf{97.4}&\textbf{72.1}\\
				&LSCM-GNN [\textcolor{blue}{TIP22}] \cite{TIP-Kan22} & 512& 68.5 & 77.3 & 85.3 & 91.3& - & \textbf{87.4} & 91.5 & 94.9 & 97.0&-\\
				\cline{2-13}
				& \textbf{Ours}: CBML &512 &\textbf{69.5} & \textbf{79.4} & \textbf{87.0} & \textbf{92.4}& \textbf{70.3} & 87.0 & \textbf{92.4} & \textbf{95.2} & 96.9& 70.7\\
				\hline
				\multirow{16}[5]*{ResNet-50}
				&Divide-Conquer [\textcolor{blue}{CVPR19}] \cite{CVPR-SanakoyeuTBO19} & 128 & 65.9 & 76.6 & 84.4 & 90.6& 69.6 & 84.6 & 90.7 & 94.1 & 96.5&70.3\\
			    &MIC+Margin [\textcolor{blue}{ICCV19}] \cite{ICCV-Roth19} & 128 &66.1 & 76.8 & 85.6 & -& 69.7& 82.6 & 89.1 & 93.2 & -&68.4\\
			    & RankMI [\textcolor{blue}{CVPR20}] \cite{CVPR20-KemertasPDF20} & 128 & 66.7 & 77.2 & 85.1 & 91.0&- & 83.3 & 89.8 & 93.8 & 96.5&-\\
			     & PADS [\textcolor{blue}{CVPR20}] \cite{CVPR20-RothMO20} & 128 & 67.3 & 78.0 & 85.9 & -&69.9 & 83.5 & 89.7 & 93.8 & -&68.8\\
			     &PA+DIML [\textcolor{blue}{\footnotesize{ICCV21}}] \cite{ICCV-ZhaoRWL021} &128 & 66.5 & - & - & -&- & 86.1 & - & - & -&-\\
			     & Sharing-DML [\textcolor{blue}{TPAMI20}] \cite{PAMI-Milbich20} & 256 & 68.6 & 79.4 & 86.8 & -&71.0 & 87.0 & 92.1 & 95.4 &-&69.8 \\
			    &TML [\textcolor{blue}{ICCV19}] \cite{ICCV-Yu2019} & 512 &62.5 & 73.9& 83.0 & 89.4&- & 86.3 & 92.3 & 95.4 & 97.3&-\\
			    & EPSHN [\textcolor{blue}{WACV20}] \cite{WACV-Xuan2020} &512&64.9 & 75.3 & 83.5 & -& -& 82.7 & 89.3 & 93.0 & -&-\\
			    & CircleLoss [\textcolor{blue}{CVPR20}] \cite{CVPR-Sun20} & 512 &66.7 & 77.4 & 86.2 & 91.2& - & 83.4 & 89.8 & 94.1 & 96.5&-\\
			    & DR [\textcolor{blue}{CVPR20}] \cite{CVPR20-MohanSFSG20} & 512&66.1 & 77.0 & 85.1 & 91.1&- & 85.0 & 90.5 & 94.1 & 96.4&-\\
			   & ProxyNCA++ [\textcolor{blue}{ECCV20}]\cite{ECCV20-TehDT20} & 512 & 69.0 & 79.8 & \textbf{87.3} & \textbf{92.7} &\textbf{73.9} & 86.5 & 92.5 & 95.7 & 97.7 & 73.8 \\
			   &DRML-MDW [\textcolor{blue}{\footnotesize{ICCV21}}] \cite{ICCV-Zheng21} & 512 & 65.7 & 76.9 & 85.6 & 91.1& 69.0&  73.3 & 83.0 & 89.8 & 94.4 & 65.3 \\
			   &DCML-MDW [\textcolor{blue}{CVPR21}] \cite{CVPR21-ZhengWL021} & 512 &68.4 & 77.9 & 86.1 & 91.7 & 71.8 & 85.2 & 91.8 & \textbf{96.0} & \textbf{98.0} & \textbf{73.9} \\
			   &MemVir+PA \textcolor{blue}{ICCV21} \cite{Ko-2021-ICCV} & 512& 69.0& - & - & - & - & 86.7 & - & - & - & - \\
			   & D \& C [\textcolor{blue}{TPAMI21}] \cite{TPAMI21-abs-2109-04003} & 512 &68.4 & 78.7 & 86.0 & 91.6 & 69.7 & 87.8 & 92.5 & 95.4 & -& 70.7\\
			   & LSCM-GNN [\textcolor{blue}{TIP22}] \cite{TIP-Kan22} & 512& 67.1 & 76.0  & 84.3 & 90.2 & - & 86.1 & 90.4 & 93.2 & 95.3 &-\\
				\cline{2-13}
				& \textbf{Ours}: CBML & 512 & \textbf{69.9} & \textbf{80.4} & 87.2 & 92.5 & 70.3 & \textbf{88.1} & \textbf{92.6} & 95.4 & 97.4 & 71.6\\
				\hline
			\end{tabularx}
			}
		\end{center}
		\label{tab:cs1}
	\end{table*}
	
	\begin{table*}[h]
		\caption{The comparison results on the SOP \cite{CVPR-SongXJS16} and the In-Shop \cite{CVPR-LiuLQWT16} datasets. It should be noted that not all methods provided results for all the evaluation metrics. In this case, they are left blank in the tables.}
		\begin{center}
		\resizebox{\linewidth}{!}{
			\begin{tabularx}{19.0cm}{c|l|c|ccccc||ccccc}
				\hline 
				\multirow{2}[1]*{Backbone} & \multirow{2}[1]*{Methods} & \multirow{2}[1]*{Dim} 	&\multicolumn{5}{c||}{\textbf{SOP}} &\multicolumn{5}{c}{\textbf{In-Shop}}\\
				\cline{4-13}
				 & & &R@1 & R@10 & R@100 & R@1000&NMI & R@1 & R@10 & R@20 & R@30&NMI \\
				\hline
				\hline
				\multirow{12}[5]*{GoogLeNet} & Contrastive [CVPR16] \cite{CVPR-SongXJS16} & 128 &42.0& 58.2& 73.8& 89.1& 82.5& - &- &- &-&-\\
				&Triplet [CVPR16] \cite{CVPR-SongXJS16} & 128 &42.1& 63.5& 82.5& 94.8& 86.3 & -& -& -& -&-\\
				& HDC [ICCV17] \cite{ICCV-YuanYZ17} & 384 &69.5& 84.4& 92.8& 97.7& -& 62.1& 84.9& 89.0& 91.2&-\\
				& LiftedStruct [CVPR16] \cite{CVPR-SongXJS16} & 512 &62.1& 79.8& 91.3& 97.4& 88.7 & -&-&-&-&- \\
				& Binomial Deviance [NIPS16] \cite{NIPS-UstinovaL16} & 512 &65.5& 82.3& 92.3& 97.6&- & -& -& -& -&-\\
				&Histogram Loss [NIPS16] \cite{NIPS-UstinovaL16} & 512 &63.9& 81.7& 92.2& 97.7&- & -&- &- &-& -\\
				& N-Pair-Loss [NIPS16] \cite{NIPS-Sohn16} & 512 &67.7& 83.8& 93.0& 97.8& 88.1 &-&-&-&-&-\\
				& Angular Loss [ICCV17] \cite{ICCV-WangZWLL17} & 512 &70.9& 85.0& 93.5& 98.0& 88.6 & - & -& -& -&-\\
				& Fusing-Net [\textcolor{blue}{TIP19}] \cite{TIP-KanCHZZW19} & 512 &71.8 & 86.3 & 94.1 & 98.2& 88.8 & 82.4 & 95.1 & 96.7 & 97.4& \textbf{88.0} \\
				& BIER [ICCV17] \cite{ICCV-OpitzWPB17} & 512 &72.7 & 86.5& 94.0& 98.0& -& 76.9& 92.8& 95.2& 96.2& -\\
				 &A-BIER [\textcolor{blue}{TPAMI20}] \cite{PAMI-Opitz18} & 512 &74.2 & 86.9 & 94.0 & 97.8& -& 83.1 & 95.1 & 96.9 & 97.5&-\\
				 &LoOp+MS[\textcolor{blue}{ICCV21}] \cite{Vasudeva-2021-ICCV} & 512 &\textbf{76.6} &\textbf{89.8} &\textbf{95.8} &- &\textbf{89.4} & - & - & - & - & - \\
				 \cline{2-13}
				 & \textbf{Ours}: CBML &512 &74.8 & 88.6 & 95.3 & \textbf{98.5}& 88.8 & \textbf{88.7} & \textbf{98.1} & \textbf{98.8} & \textbf{99.1}& \textbf{88.0}\\
				 \hline
				\multirow{11}[4]*{BN-Inception} & RLL-Simpler [\textcolor{blue}{TPAMI21}] \cite{TPAMI21-Wang} & 384 &79.3 & \textbf{91.3} & \textbf{96.3} & - & -& 89.9 & 97.6 & 98.3 & 98.7&-\\
				&RLL-H [\textcolor{blue}{CVPR19}] \cite{CVPR-WangHKHGR19} & 512&76.1 & 89.1 & 95.4 &  -&89.7 & - & - & - & -&-\\
				&RLL-(L,M,H) [\textcolor{blue}{CVPR19}] \cite{CVPR-WangHKHGR19} &1536 &79.8 & \textbf{91.3} & \textbf{96.3} & -& 90.4& - & - & - & -&-\\
				&SoftTriple [\textcolor{blue}{ICCV19}] \cite{ICCV-Qian2019} &512 & 78.3 & 90.3 & 95.9 & - & \textbf{92.0} & - & - & - & -& -\\
				&DeML [\textcolor{blue}{ICCV19}] \cite{CVPR-ChenD19} &512 &76.1 & 88.4 & 94.9 & 98.1& -& 88.2 & 97.0 & 98.0 & 98.3&-\\
				&MS [\textcolor{blue}{CVPR19}] \cite{CVPR-WangHHDS19} &512 &78.2 & 90.5 & 96.0 & \textbf{98.7}& - & 89.7 & 97.9 & 98.5 & 98.8&-\\
				&Contrastive+HORDE [\textcolor{blue}{ICCV19}] \cite{ICCV-Jacob2019} &512 &\textbf{80.1} & \textbf{91.3} & 96.2 & \textbf{98.7}&- & 90.4 & 97.8 & 98.4 & 98.7& -\\
				&MS+XBM [\textcolor{blue}{CVPR20}] \cite{CVPR-Wang20}& 512&79.5 & 90.8 & 96.1 & \textbf{98.7}& - & 89.9 & 97.6 & 98.4 & 98.6&- \\
				& Proxy-Anchor [\textcolor{blue}{CVPR20}]\cite{CVPR20-KimKCK20} & 512 & 79.1 & 90.8 & 96.2 & \textbf{98.7}& -& 91.5 & 98.1 & 98.8 & 99.1& -\\ 
				&LSCM-GNN [\textcolor{blue}{TIP22}] \cite{TIP-Kan22} & 512& 79.7 & 90.5 & 95.7 & 98.4& -& \textbf{92.4} & 98.5 & 99.1 & 99.3&-\\
				\cline{2-13}
				& \textbf{Ours}: CBML &512& 77.5 & 90.0 & 95.9 &\textbf{98.7} & 89.1 & 91.7 & \textbf{98.8} & \textbf{99.4} & \textbf{99.5}& \textbf{89.0} \\
				\hline
				\multirow{17}[5]*{ResNet-50}&Margin [ICCV17] \cite{ICCV-ManmathaWSK17} & 128 &72.7 & 86.2& 93.8& 98.0& 90.7& - &- &- &-&- \\
				&Divide-Conquer [\textcolor{blue}{CVPR19}] \cite{CVPR-SanakoyeuTBO19} & 128& 75.9 & 88.4 & 94.9 & 98.1& 90.2& 85.7 & 95.5 & 96.9 & 97.5&88.6\\
				&MIC+Margin [\textcolor{blue}{ICCV19}] \cite{ICCV-Roth19} & 128 &77.2 & 89.4 & 95.6 & -& 90.0 & 88.2 & 97.0 & 98.0 & 98.8& -\\
				 & PADS [\textcolor{blue}{CVPR20}] \cite{CVPR20-RothMO20} & 128 &76.5 & 89.0 & 95.4 & - &89.9 & - & - & - & -&- \\
				 & RLL-Simpler [\textcolor{blue}{TPAMI21}] \cite{TPAMI21-Wang} & 128 &78.7 & 91.1 & 96.4 & -&- & - & - & - & -&-\\
				 &PA+DIML [\textcolor{blue}{\footnotesize{ICCV21}}] \cite{ICCV-ZhaoRWL021}&128 & 79.2 & - & - & -& - & - & - & - & -&-\\
			      & Sharing-DML [\textcolor{blue}{TPAMI20}] \cite{PAMI-Milbich20} & 256 &78.2 & 90.1 & 96.1 &- &90.3 & - & - & - &-&- \\
			    &FastAP [\textcolor{blue}{CVPR19}] \cite{CVPR-Cakir0XKS19} &512 &76.4 & 89.0 & 95.1 & 98.2 & -& 90.9 & 97.7 & 98.5 & 98.8&- \\
			    &TML [\textcolor{blue}{ICCV19}] \cite{ICCV-Yu2019} &512 &78.0 & 91.2 & \textbf{96.7} & \textbf{99.0} &- & - & - & - & -&-\\
			    & EPSHN [\textcolor{blue}{WACV20}] \cite{WACV-Xuan2020} & 512 &78.3 & 90.7 & 96.3 & -&- & 87.8 & 95.7 &  96.8 & -&- \\
			   & CE [\textcolor{blue}{\footnotesize{ECCV20}}] \cite{ECCV-BoudiafRZGPPA20}&2048 & \textbf{81.1} & 91.7 & 96.3 & 98.8& - & 90.6 & 98.0 & 98.9 & 99.1&- \\
			    & CircleLoss [\textcolor{blue}{CVPR20}] \cite{CVPR-Sun20} & 512 &78.3 & 90.5 & 96.1 & 98.6& - & - & - & - & -& -\\
			    & ProxyNCA++ [\textcolor{blue}{ECCV20}] \cite{ECCV20-TehDT20} & 512 & 80.7 & \textbf{92.0} & \textbf{96.7} & 98.9& -& 90.4 & 98.1 & 98.8 & 99.0& -\\
			   & DRML-MDW [\textcolor{blue}{\footnotesize{ICCV21}}] \cite{ICCV-Zheng21}& 512 & 79.9 & 90.7 & 96.1 & -& 90.1 &  - & - & - & -&- \\
			    &DCML-MDW [\textcolor{blue}{CVPR21}] \cite{CVPR21-ZhengWL021} & 512&79.8 & 90.8 & 95.8 & -&\textbf{90.8} &-&-&-&-& -\\
			    &MemVir+PA \textcolor{blue}{ICCV21} \cite{Ko-2021-ICCV} & 512& 79.7& - & - & - & - & - & - & - & - & - \\
			    & D \& C [\textcolor{blue}{TPAMI21}] \cite{TPAMI21-abs-2109-04003} & 512 &79.8 & 90.4 & 95.2 & -& 89.7 & 90.4 & 97.6 & - & -& \textbf{89.9}\\
			    &LSCM-GNN [\textcolor{blue}{TIP22}] \cite{TIP-Kan22} & 512& 80.5 & 89.1 & 94.2 & 97.8& -& 91.9 & 97.5 & 98.1 & 98.4&- \\
			    \cline{2-13}
				& \textbf{Ours}: CBML & 512& 79.9 & 91.5 & 96.5 & 98.9& 90.1 & \textbf{92.3} & \textbf{98.7} & \textbf{99.2} & \textbf{99.4}& 89.4 \\
				\hline
			\end{tabularx}
			}
		\end{center}
		\label{tab:cs2}
	\end{table*}

\subsection{Performance Metrics and Experimental Settings}
We follow the  standard evaluation protocol \cite{CVPR-SongXJS16,PAMI-Opitz18,TIP-KanCHZZW19,PAMI-Milbich20} and use the Recall@K \cite{PAMI-JegouDS11}  to evaluate the performance of our algorithm. For all datasets, we only use the original images without the object bounding box information.
We apply random cropping and flipping to all training images and resize them  to 227 $\times$ 227. For testing, we only use the  center-cropped images to compute the feature embedding. To compare with the state-of-the-art methods, we use GoogLeNet \cite{CVPR-SzegedyLJSRAEVR15}, BN-Inception \cite{ICML-IoffeS15}, and ResNet-50 \cite{CVPR-HeZRS16} with an one-layer embedding head to embed the representation to the 512-dimensional feature  space. To verify the generalization performance of our algorithm, we further compare it with the state-of-the-art methods on 64-dimensional feature embedding. For fair comparison, we use the same GoogLeNet \cite{CVPR-SzegedyLJSRAEVR15}, ResNet-18, and ResNet-50 \cite{CVPR-HeZRS16} with an 1-layer embedding head to embed the representation to 64-dimensional feature embedding space.

We implement our algorithm with PyTorch on one GeForce GTX 1080 GPU with 11GB memory, and use the Adam optimization algorithm \cite{ICLR-KingmaB14} in all  experiments.  For all images, the parameters of $\lambda$ is set as 1.0 for CUB and Cars datasets, and set as 0.001 for SOP and In-Shop datasets. For all datasets, $\delta^P$ and $\delta^N$ are set as 1. For the CUB and Cars datasets, $\alpha^P$, $\beta^P$, $\alpha^N$, and $\beta^N$ are set as 0.5, 0.5, 1.0, and 0.01, respectively. For SOP and In-Shop datasets, $\alpha^P$, $\beta^P$, $\alpha^N$, and $\beta^N$ are set as 0.5, 0.25, 0.5, and 0.05, respectively. The settings of these parameters are analyzed in ablation studies.  The backbone network is pre-trained on the ImageNet-1K dataset. For datasets for fine-grained object classification, \cite{CVPR-WangHHDS19} and \cite{CVPR-Wang20} have shown that fixing the parameters of the batch normalization (BN) layer can enhance the generalization ability from the training set to the test set. We also freeze the parameters of the BN layer in our experiments.

\subsection{Performance Comparisons with the State-of-the-Art Methods}

We compare our method with the following state-of-the-art methods recently developed in the literature: \textit{Fusing-Net} \cite{TIP-KanCHZZW19}, \textit{A-BIER} \cite{PAMI-Opitz18}, \textit{DeML} \cite{CVPR-ChenD19},\textit{ RLL} \cite{CVPR-WangHKHGR19}, \textit{MS} \cite{CVPR-WangHHDS19}, \textit{HORDE} \cite{ICCV-Jacob2019}, \textit{SoftTriple} \cite{ICCV-Qian2019}, \textit{MS+EE} \cite{cvpr20-KoG20}, \textit{MS+XBM} \cite{CVPR-Wang20}, \textit{Proxy-Anchor} \cite{CVPR20-KimKCK20}, \textit{DRML-PA} \cite{ICCV-Zheng21}, \textit{LoOp+MS} \cite{Vasudeva-2021-ICCV}, and \textit{LSCM-GNN} \cite{TIP-Kan22}. These methods are based on the GoogLeNet \cite{CVPR-SzegedyLJSRAEVR15} or GoogLeNet with  batch normalization \cite{ICML-IoffeS15} (BN-Inception) backbone networks. We also compare our method with \textit{Divide-Conquer} \cite{CVPR-SanakoyeuTBO19}, \textit{FastAP} \cite{CVPR-Cakir0XKS19}, \textit{MIC+Margin} \cite{ICCV-Roth19}, \textit{TML} \cite{ICCV-Yu2019}, EPSHN \cite{WACV-Xuan2020}, \textit{CircleLoss} \cite{CVPR-Sun20}, \textit{Sharing-DML} \cite{PAMI-Milbich20}, \textit{DR} \cite{CVPR20-MohanSFSG20}, \textit{RankMI} \cite{CVPR20-KemertasPDF20}, \textit{PADS} \cite{CVPR20-RothMO20}, \textit{DCML-MDW} \cite{CVPR21-ZhengWL021}, \textit{RLL-Simpler} \cite{TPAMI21-Wang}, \textit{D \& C} \cite{TPAMI21-abs-2109-04003}, \textit{PA+DIML} \cite{ICCV-ZhaoRWL021}, \textit{CircleLoss} \cite{CVPR-Sun20}, \textit{ProxyNCA++} \cite{ECCV20-TehDT20}, \textit{DRML-MDW} \cite{ICCV-Zheng21}, \textit{MemVir+PA} \cite{Ko-2021-ICCV}, and \textit{LSCM-GNN} \cite{TIP-Kan22}, which are based on the ResNet-50 \cite{CVPR-HeZRS16} backbone network. A brief review of these algorithms are provided in the Related Work of Section \ref{sec:relatedwork}. It should be noted that not all methods have provided results for all four datasets. If they are not available in the original papers, they are left blank in the following tables.  

\begin{table*}[h]
		\caption{Comparisons of retrieval performance on the CUB and Cars datasets with different backbones based on 64 dimensional feature embedding size. It should be noted that not all methods provided results for all the backbone networks. In this case, they are left blank in the tables.}
		\begin{center}
		\resizebox{\linewidth}{!}{
			\begin{tabularx}{17.5cm}{c|l|cccc||cccc||cccc}
				\hline 
				 \multirow{2}[1]*{Datasets} & \multirow{2}[1]*{Methods}	&\multicolumn{4}{c||}{\textbf{GoogLeNet}} &\multicolumn{4}{c||}{\textbf{ResNet-18}}&\multicolumn{4}{c}{\textbf{ResNet-50}}\\
				\cline{3-14} 
				&& R@1 & R@2 & R@4 & R@8 & R@1 & R@2 & R@4 & R@8 & R@1 & R@2 & R@4 & R@8 \\
				\hline
				\hline
				\multirow{7}[4]*{CUB}
				&Triplet [CVPR15] \cite{CVPR-SchroffKP15} &42.6 & 55.0  & 66.4  & 77.2 & - & - &- &- &- &-&-&-\\
				&N-Pair [NIPS16] \cite{NIPS-Sohn16} & 51.0& 63.3& 74.3& 83.2 & 52.4 &  65.7 &  76.8 &  84.6 &53.2  & 65.3 &  76.0  & 84.8\\
				&ProxyNCA [ICCV17] \cite{ICCV-Movshovitz-Attias17} & 49.2 &  61.9 &  67.9 &  72.4& 51.5 &  63.8 &  74.6 &  84.0& 55.5  & 67.7 &  78.2  & 86.2 \\
				& Clustering [CVPR17] \cite{CVPR-SongJR017} & 48.2& 61.4& 71.8& 81.9 & - & - &- &- &- &-&-&-  \\
				& Smart Mining [ICCV17] \cite{ICCV-HarwoodGCRD17} & 49.8& 62.3& 74.1& 83.3& - & - &- &- &- &-&-&-\\
				&EPSHN [\textcolor{blue}{WACV20}] \cite{WACV-Xuan2020} & 51.7 &  64.1 &  75.3 & 83.9 & 54.2 & 66.6  & 77.4 &  86.0 & 57.3  & 68.9  & 79.3  & 87.2 \\
				& SCT [\textcolor{blue}{ECCV20}] \cite{ECCV20-XuanSLP20} & -&-&-&-&-&-&-&-& \textcolor{blue}{57.7} & \textcolor{blue}{69.8} & 79.6 & 87.0 \\
				&MS [\textcolor{blue}{CVPR19}] \cite{CVPR-WangHHDS19} & \textcolor{blue}{54.4}&\textcolor{blue}{66.2}&\textcolor{blue}{77.1}&\textcolor{blue}{86.1}&\textcolor{blue}{55.0}&\textcolor{blue}{67.6}&\textcolor{blue}{77.9}&\textcolor{blue}{86.2}& 57.4 & \textcolor{blue}{69.8} & \textcolor{blue}{80.0} & \textcolor{blue}{87.8} \\
				\cline{2-14}
				&\textbf{Ours}: CBML & \textbf{59.3} & \textbf{70.7} & \textbf{80.6} & \textbf{88.1} & \textbf{61.3} & \textbf{72.6} & \textbf{81.9} & \textbf{88.7} & \textbf{64.3} & \textbf{75.7} & \textbf{84.1} & \textbf{90.1} \\
				&\textbf{Gain} & \textcolor{red}{\textbf{+4.9}}&\textcolor{red}{\textbf{+4.5}} & \textcolor{red}{\textbf{+3.5}}&\textcolor{red}{\textbf{+2.0}}&\textcolor{red}{\textbf{+6.3}}&\textcolor{red}{\textbf{+5.0}}&\textcolor{red}{\textbf{+4.0}}&\textcolor{red}{\textbf{+2.5}} & \textcolor{red}{\textbf{+6.6}}&\textcolor{red}{\textbf{+5.9}}&\textcolor{red}{\textbf{+4.1}}&\textcolor{red}{\textbf{+2.3}}\\
				\hline
				\hline
				\multirow{7}[3]*{Cars}
				&Triplet [CVPR15] \cite{CVPR-SchroffKP15} &51.5 & 63.8  & 73.5  & 81.4 & - & - &- &- &- &-&-&-\\
				&N-Pair [NIPS16] \cite{NIPS-Sohn16} & 71.1& 79.7&86.5& 91.6 & 55.7  & 67.4  & 77.0 &  84.5 & 58.3 &  69.5  & 78.3 &  86.4\\
				&ProxyNCA [ICCV17] \cite{ICCV-Movshovitz-Attias17} & 73.2 &  82.4 &  86.4 & 88.7 & 62.2 &  73.0 &  81.6  & 87.9 & 66.2 &  76.9 &  84.9  & 90.5 \\
				& Clustering [CVPR17] \cite{CVPR-SongJR017}& 58.1& 70.6& 80.3& 87.8& - & - &- &- &- &-&-&- \\
				& Smart Mining [ICCV17] \cite{ICCV-HarwoodGCRD17} & 64.7& 76.2& 84.2& 90.2& - & - &- &- &- &-&-&-\\
				&EPSHN [\textcolor{blue}{WACV20}] \cite{WACV-Xuan2020} & 66.4  & 76.8 &  85.2 & 90.7 & \textcolor{blue}{73.2} &  \textcolor{blue}{82.5}  & \textcolor{blue}{88.6} &  93.0 & 75.5  & 84.2 &  90.3  & \textcolor{blue}{94.2} \\
				& SCT [\textcolor{blue}{ECCV20}] \cite{ECCV20-XuanSLP20} &-&-&-&-&-&-&-&-& 73.4 & 82.0 & 88.0 & 92.4 \\
				&MS [\textcolor{blue}{CVPR19}] \cite{CVPR-WangHHDS19} & \textcolor{blue}{76.5}&\textcolor{blue}{84.4}&\textcolor{blue}{89.8}&\textcolor{blue}{93.9}&72.3&82.0&88.1&\textcolor{blue}{93.2}& \textcolor{blue}{77.3} & \textcolor{blue}{85.3} & \textcolor{blue}{90.5} & \textcolor{blue}{94.2} \\
				\cline{2-14}
				&\textbf{Ours}: CBML & \textbf{77.0} & \textbf{85.0} & \textbf{90.5} & \textbf{94.2} & \textbf{77.6} & \textbf{85.8} & \textbf{90.8} & \textbf{94.5} & \textbf{83.2} & \textbf{89.4} & \textbf{93.5} & \textbf{96.0}\\
				&\textbf{Gain} & \textcolor{red}{\textbf{+0.5}}&\textcolor{red}{\textbf{+0.6}} & \textcolor{red}{\textbf{+0.7}}&\textcolor{red}{\textbf{+0.3}}&\textcolor{red}{\textbf{+4.4}}&\textcolor{red}{\textbf{+3.3}}&\textcolor{red}{\textbf{+2.2}}&\textcolor{red}{\textbf{+1.3}} & \textcolor{red}{\textbf{+5.9}}&\textcolor{red}{\textbf{+4.1}}&\textcolor{red}{\textbf{+3.0}}&\textcolor{red}{\textbf{+1.8}}\\
				\hline
			\end{tabularx}
			}
		\end{center}
		\label{tab:cs3}
	\end{table*}

\begin{table*}[h]
		\caption{Impact of different averaging operations on CUB, Cars, SOP, and In-Shop datasets for GoogLeNet, BN-Inception and ResNet-50 backbones.}
		\begin{center}
		\resizebox{\linewidth}{!}{
			\begin{tabularx}{21.6cm}{c|l|cccc||cccc||cccc||cccc}
				\hline 
				\multirow{2}[1]*{Backbone} & \multirow{2}[1]*{Methods}	&\multicolumn{4}{c||}{\textbf{CUB}} &\multicolumn{4}{c||}{\textbf{Cars}}&\multicolumn{4}{c||}{\textbf{SOP}}&\multicolumn{4}{c}{\textbf{In-Shop}}\\
				\cline{3-18} 
				& & R@1 & R@2 & R@4 & R@8 & R@1 & R@2 & R@4 & R@8 & R@1 & R@10 & R@100 & R@1000&R@1 & R@10 & R@20 & R@30 \\
				\hline
				\hline
				\multirow{3}[1]*{GoogLeNet} 
				& CBML-const&62.8& 73.9& 83.2&89.8&79.4 & 87.3 & 92.3 & 95.3  & 70.3 & 85.4 & 93.8 & 98.1 & 80.4 & 95.7 & 97.4 & 98.0 \\
				& CBML-sqrt & 63.1 & 74.7 & 83.1 & 89.8 & 81.6 & 88.3 & 92.9 & 95.8 & 74.4 & 88.4 & 95.2 & \textbf{98.5} & 87.7 & 97.8 & 98.7 & \textbf{99.1} \\
				& CBML & \textbf{63.8} & \textbf{74.8} & \textbf{83.6} & \textbf{90.3} & \textbf{83.0} & \textbf{89.3} & \textbf{93.5} & \textbf{96.2} & \textbf{74.8} & \textbf{88.6} & \textbf{95.3} & \textbf{98.5} & \textbf{88.7} & \textbf{98.1} & \textbf{98.8} & \textbf{99.1}\\
				\hline
				\multirow{3}[1]*{BN-Inception}
				& CBML-const&68.3& 78.5& 86.9&92.1&84.5& 90.7& 94.5 & \textbf{96.9} & 73.6 & 87.6 & 94.7 & 98.3 & 88.4 & 98.1 & 98.9 & 99.2 \\
				& CBML-sqrt & \textbf{69.5} & \textbf{79.5} & 86.7 & 91.8 & 85.4 & 90.8& 94.4 & \textbf{96.9} & 76.5 & 89.8 & 95.8 & \textbf{98.7} & 90.2 & 98.5 & 99.2 & \textbf{99.5}\\
				& CBML & \textbf{69.5} & 79.4 & \textbf{87.0} & \textbf{92.4} & \textbf{87.0} & \textbf{92.4} & \textbf{95.2} & \textbf{96.9} & \textbf{77.5} & \textbf{90.0} & \textbf{95.9} &\textbf{98.7} & \textbf{91.0} & \textbf{98.8} & \textbf{99.4} & \textbf{99.5}\\
				\hline
				\multirow{3}[1]*{ResNet-50}
			   & CBML-const&69.2& 79.3& 86.3&91.6& 86.1 & 91.8 & 95.3& \textbf{97.5} & 74.2 & 87.5 & 94.4 & 98.0 & 84.7 & 97.1 & 98.3 & 98.8\\
				& CBML-sqrt & \textbf{70.0} & 79.9 & 87.0 &92.0 & 87.3 & \textbf{92.7} & \textbf{95.6} & \textbf{97.5} & 78.6 & 90.7 & 96.1 & 98.8 & 89.7 & 98.4 & 99.1 & 99.3\\
				& CBML & 69.9 & \textbf{80.4} & \textbf{87.2} & \textbf{92.5} & \textbf{88.1} & 92.6 & 95.4 & 97.4  & \textbf{79.9} & \textbf{91.5} & \textbf{96.5} & \textbf{98.9} & \textbf{92.3} & \textbf{98.7} & \textbf{99.2} & \textbf{99.4} \\
				\hline
			\end{tabularx}
			}
		\end{center}
		\label{tab:ab-average}
	\end{table*}

The performance comparisons with existing methods on the CUB and Cars datasets are summarized in Table \ref{tab:cs1}.
The results on the SOP and In-Shop datasets are summarized in Table \ref{tab:cs2}. 
From Table \ref{tab:cs1} and Table \ref{tab:cs2}, we can see that our method outperforms the state-of-the-art methods by a large margin, especially on the CUB dataset. 
With the same GoogLeNet backbone encoder, we have improved the state-of-the-art top-1 recall rates by 1.9\%, 1.0\% and 5.6\% on the CUB, Cars, and In-Shop datasets, respectively. With the BN-Inception backbone, we have improved the state-of-the art top 1 recall rates by 0.8\% on the CUB dataset. With the ResNet-50 backbone encoder, we have improved the state-of-the art top-1 recall rates by 0.9\% and 0.3\% on the CUB and Cars datasets, respectively. It should be noted that we only use one GPU with 11GB memory to conduct our experiments. Our results on the SOP dataset are using a batch-size of 200 for GoogLeNet and BN-Inception, and 100 for ResNet-50. It should be noted that the results on the SOP dataset by the MS method were obtained by using a very large batch size of 1000. Moreover, the LSCM-GNN uses graph to merge local information to improve the top-1 recall rates, which is complementary with our CBML method.

We do notice that, on the CUB and In-Shop datasets, the top-2 recall rates of our CBML method have achieved very considerable improvement, 1.9\%, 0.2\%, and 0.6\% for GoogLeNet, Bn-Inception, and ResNet-50 backbones on the CUB dataset, 3.0\%, 0.3\%, and 0.6\% for GoogLeNet, Bn-Inception, and ResNet-50 backbones on the In-Shop dataset, respectively. Overall, the performance improvement achieved by our method is significant.

Moreover, we can see that the NMI scores are not consistent or highly correlated with the actual embedding performance measured by the retrieval accuracy. This is because the NMI is a measure for the clustering performance at the feature space. The retrieval accuracy is a measurement at the semantic label space.

The performance comparisons on the CUB and Cars datasets with GoogLeNet, ResNet-18 and ResNet-50 backbone networks on the 64 dimensional feature embeddings are shown in Table \ref{tab:cs3}.
We can see that our method outperforms the state-of-the-art methods by a large margin. On the CUB dataset, using the GoogLeNet, ResNet-18, and ResNet-50 backbone networks, our CBML method has improved the top-1 recall rates by 4.9\%, 6.3\% and 6.6\%, respectively. On the Cars dataset, it improves the Recall@1 rates by 0.5\%, 4.4\% and 5.9\%, respectively.

\begin{figure*}[t]
	\begin{center}
    \includegraphics[width=0.9\linewidth]{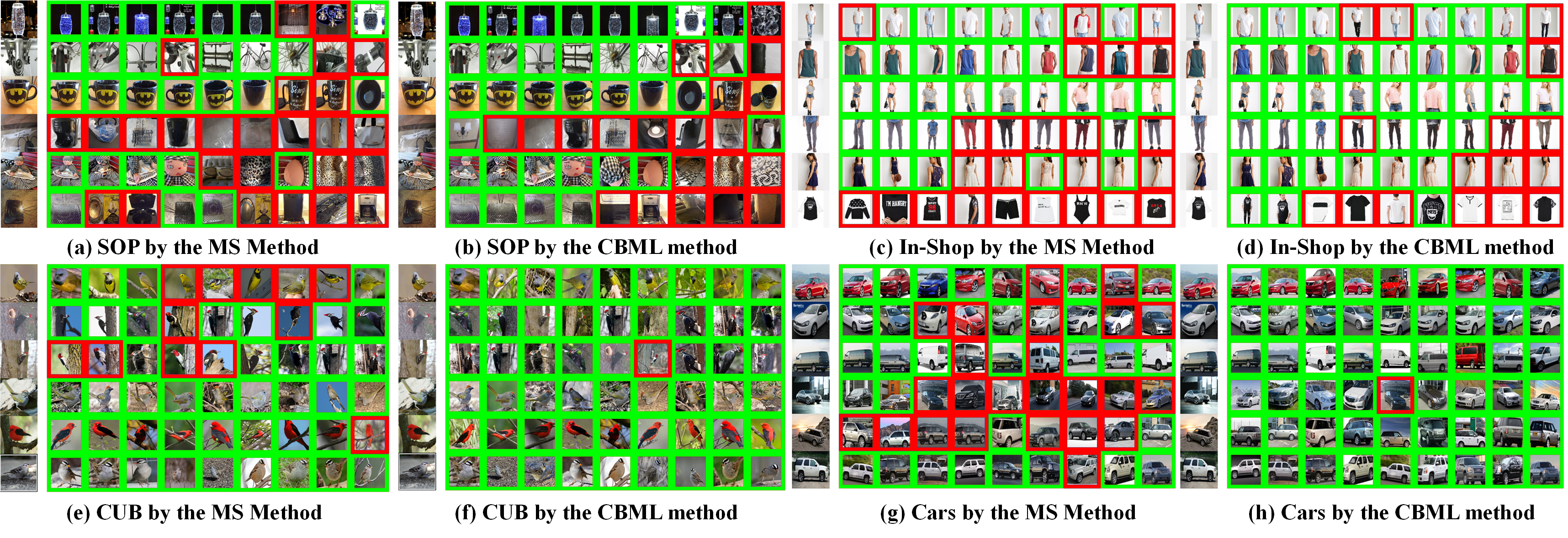}
    \end{center}
    \vspace{-4mm}
	\caption{Retrieval examples by the MS method \cite{CVPR-WangHHDS19} and our CBML method on the SOP, In-Shop, CUB and Cars datasets based on the ResNet-50 backbone. (a) and (b) are results on the SOP dataset, (c) and (d) are results on the In-Shop dataset, (e) and (f) are results on the CUB dataset, (g) and (h) are results on the Cars dataset. Retrieved images with green boxes are correct ones with the same class label as the query image. Those with red boxes are incorrect results from other classes. }
    \label{fig:examples}
    \vspace{-4mm}
\end{figure*}

Fig. \ref{fig:examples} shows examples of retrieval results by the MS method \cite{CVPR-WangHHDS19} and our CBML method on the  SOP, In-Shop, CUB, and Cars datasets. We select the MS method for comparison because it provides the original source code for the algorithm. 
In each row, the first image is the query image, followed by the retrieval results. Retrieved images with green boxes are correct ones with the same class label as the query image. Those with red boxes are incorrect results from other classes. We can see that using our CBML method for feature embedding, the retrieval system returns many more correct results (more green boxes) in the top matches than the MS method \cite{CVPR-WangHKHGR19}.

\subsection{Ablation Studies}

In this section, we conduct ablation studies on the benchmark datasets to further understand the performance of our CBML methods.

\textbf{(A) Impact of different averaging operations}
As discussed in Section \ref{sec-averaging}, there are three different ways to perform the average operation on the conditional probabilities or three different options for the function $f(\cdot)$
in (\ref{eq-avg-options}): $f(x) = \log(x)$, $f(x) = x$, and $f(x) = \sqrt{x}$, and the corresponding algorithms are denoted by CBML, CBML-const, and CBML-sqrt. Here, our default CBML algorithm is using $f(x)=\log(x)$ for the averaging operation. The performance of these three different averaging operations on CUB, Cars, SOP and In-Shop datasets are shown in Table \ref{tab:ab-average}.
We can see that, on these four datasets, the CBML method is better than the CBML-const and  CBML-sqrt methods. This is why we have chosen the logarithm function for the averaging operation in (\ref{eq-org-loss-1}).

\textbf{(B) Impact of the embedding size on the CBML performance.} The dimension of the feature embedding has a direct impact on the retrieval accuracy. We conduct experiments with different dimensions of embedding on the CUB, Cars, and In-Shop datasets. Results with the ResNet-50 backbone are shown in Table \ref{tab:as1}. We can see that the best embedding performance is achieved at size of 512 for CUB and Cars datasets, and 1024 for In-Shop dataset. It shows that, at the dimension of feature embedding is 512, the retrieval performance  saturates.

\begin{table*}[t]
		\caption{The results with ResNet-50 backbone for different dimensions of embedding on the CUB, Cars and In-Shop datasets with CBML.}
		\begin{center}
		\resizebox{0.85\linewidth}{!}{
			\begin{tabularx}{13.8cm}{c|cccc||cccc||cccc}
				\hline 
				 \multirow{2}[1]*{Dim}	&\multicolumn{4}{c||}{\textbf{CUB}} &\multicolumn{4}{c||}{\textbf{Cars}}&\multicolumn{4}{c}{\textbf{In-Shop}}\\
				\cline{2-13} 
				& R@1 & R@2 & R@4 & R@8 & R@1 & R@2 & R@4 & R@8 & R@1 & R@10 & R@20 & R@30 \\
				\hline
				\hline
				64 & 64.3 & 75.7 & 84.1 & 90.1 & 83.2 & 89.4 & 93.5 & 96.0 & 88.9 & 97.8 & 98.7 & 99.1\\
				128 & 66.9 & 77.7 & 85.2 & 90.8 & 86.5 & 91.7 & 95.3 & 97.1 & 90.9 & 98.4 & 99.0 & 99.3 \\
				256 & 68.6 & 78.4 & 86.3 & 91.5 & 86.8 & 91.8 & \textbf{95.4} & \textbf{97.4} & 90.9 & 98.7 & \textbf{99.3} & \textbf{99.4}\\
				512 & \textbf{69.9} & \textbf{80.4} & \textbf{87.2} & \textbf{92.5} & \textbf{88.1} & \textbf{92.6} & \textbf{95.4} & \textbf{97.4} & 92.3 & 98.7 & 99.2 & \textbf{99.4} \\
				1024& 69.0 & 78.7 & 86.1 & 91.2 & 86.9 & 91.8 & 94.9 & 96.9 & \textbf{92.5} & \textbf{98.8} &  \textbf{99.3} & \textbf{99.4}\\ 
				\hline
			\end{tabularx}
			}
		\end{center}
		\label{tab:as1}
		\vspace{-4mm}
	\end{table*}

\textbf{(C) Impact of the metric variance constraint on the generalization and convergence.} As discussed in our method, the metric variance constraint of the proposed CBML method can improve the generalization performance of the trained model. To demonstrate this capability, we compute the top-1 recall rates on the training and test sets of CUB dataset using the ResNet-50 backbone. Results are shown in Table \ref{tab:ab-distribution-constraint}, we can see that, as $\lambda$ increases, the top-1 recall rates are increased on the test set.

From Table \ref{tab:ab-distribution-constraint}, we can see that the top-1 recall rates on the test set with ($\lambda=1.0$) and without ($\lambda=0.0$) the variance constraint (\ref{eq:constraint}) are 69.9\% and 63.8\%, respectively. This shows the effectiveness of the metric variance constraint on the test set. The top-1 recall rates for $\lambda=1.0$ and $\lambda=0.0$ are 68.6\% and 70.1\% on the training set, which shows the over-fitting problem with loss function (\ref{eq:log-bayes-transform}) on the training set. Because our CBML method directly optimizes the global distribution of metric distance, it will overfit the embedding distribution of the training set if the MVC is not used (i.e., lambda is 0.0). In our experiments, the test classes are different from the training classes. When the training set is overfit, the model cannot generalize well on the test set. By constraining the distribution of the embedding, the overfitting problem can be mitigated. 

\begin{table}[t]
		\caption{The Recall@1 rates on the CUB training and test sets for different weights of $\lambda$ based on the ResNet-50 backbone and CBML method. The first row is the results on the training set, the second row is the results on the test set..}
		\begin{center}
		\resizebox{0.7\linewidth}{!}{
			\begin{tabularx}{5.5cm}{l|cccc}
				\hline 
				 $\lambda$ & 0.0 & 0.1 & 1.0 & 2.0 \\
				\hline
				\hline
				Training Set &70.1 & 71.9 & 68.6 & 68.7\\
				Test Set &63.8 & 65.9 & \textbf{69.9} & 69.6\\
				\hline
			\end{tabularx}
			}
		\end{center}
		\label{tab:ab-distribution-constraint}
		\vspace{-4mm}
	\end{table}

\begin{figure}[t]
    \begin{center}
        \includegraphics[width=\linewidth]{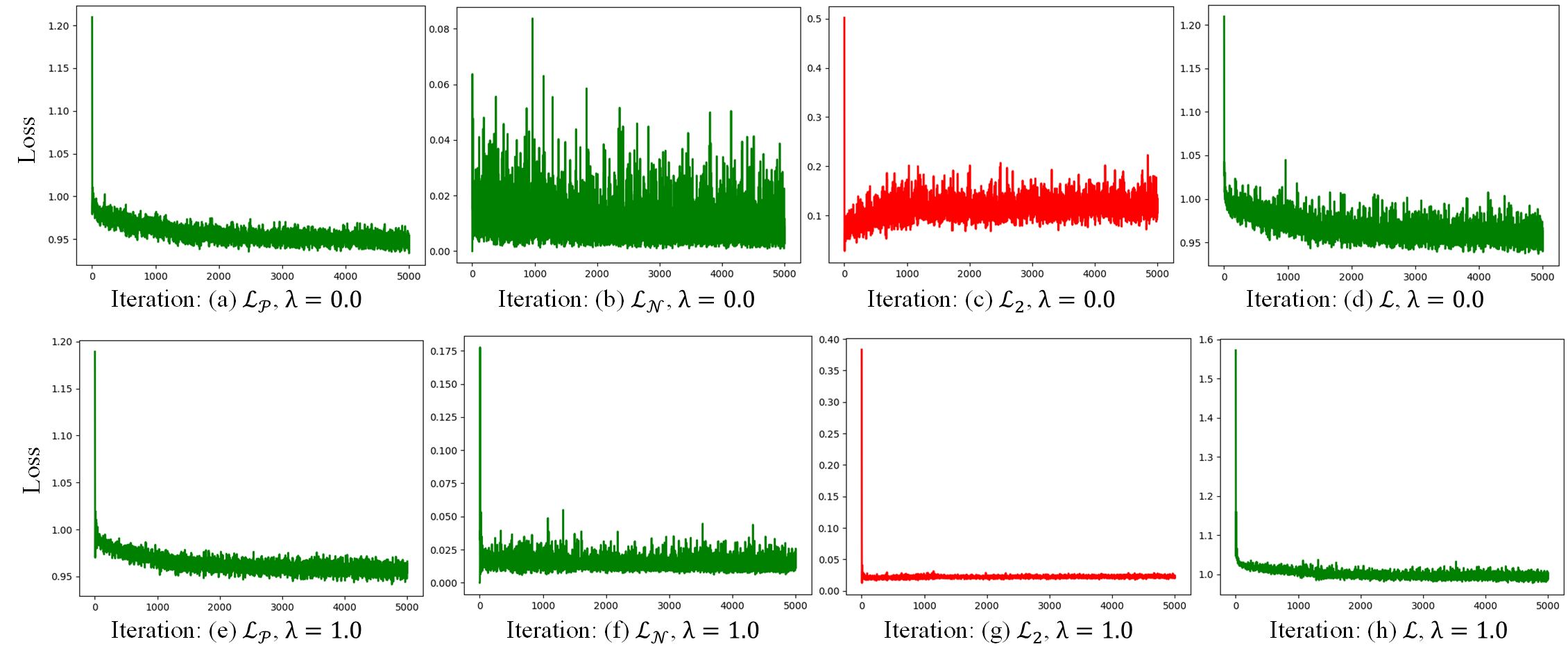}
    \end{center}
    \vspace{-4mm}
	\caption{The loss value change of CBML method on the CUB dataset based on the ResNet-50 backbone for the loss of positive pairs $\mathcal{L}_{\mathcal{P}}$, the loss of negative pairs $\mathcal{L}_{\mathcal{N}}$, loss $\mathcal{L}_2$, and loss $\mathcal{L}$. Figures in the first row are the results with $\lambda=0.0$, and figures in the second row are the results with $\lambda=1.0$.}
	\label{fig:loss-cub}
\end{figure}

To understand the convergence of the proposed method with and without the metric variance constraint (\ref{eq:constraint}), in Fig. \ref{fig:loss-cub}, we show the convergence behaviors of different components of the loss function, specifically, $\mathcal{L}_{\mathcal{P}}$, $\mathcal{L}_{\mathcal{N}}$, $\mathcal{L}_2$, and $\mathcal{L}$, on the CUB  dataset with the ResNet-50 backbone and the CBML loss function. 
Here, $\mathcal{L}_{\mathcal{P}}$ and $\mathcal{L}_{\mathcal{N}}$ are the losses for positive and negative pairs, respectively. They are very important for the deep metric learning process. When their values are decreasing, the learned features are more effective.
Compared to Figs. \ref{fig:loss-cub}(c) and \ref{fig:loss-cub}(g), we can see that the $\mathcal{L}_2$ loss gradually increases with the  training iteration if the constraint (\ref{eq:constraint}) is not used. However, if this constraint is used, this $\mathcal{L}_2$ loss converges quickly after a few iterations. Compared to the overall loss values shown in Figs. \ref{fig:loss-cub} (d) and (h), the loss values without constraint (\ref{eq:constraint}) are much smaller than those  with the constraint (\ref{eq:constraint}). This indicates that it is easier to over-fit on the training set when there is no constraint. 

\begin{figure}[t]
    \begin{center}
        \includegraphics[width=\linewidth]{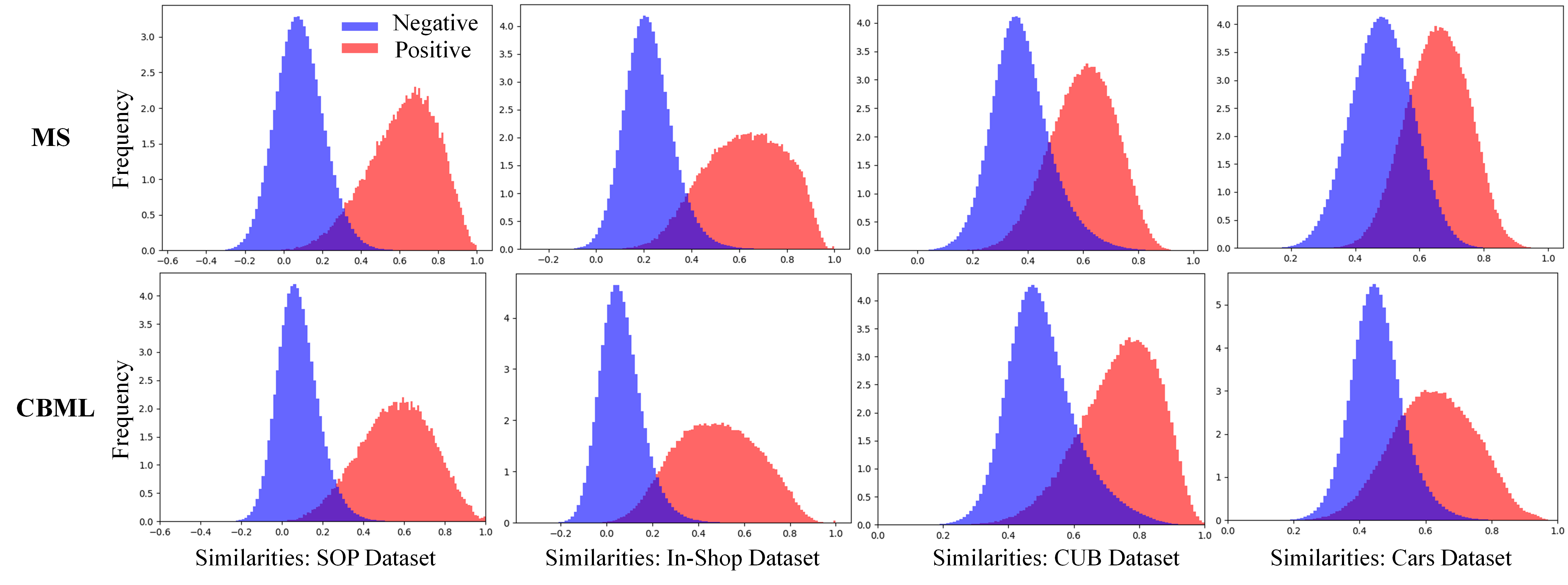}
    \end{center}
    \vspace{-4mm}
	\caption{The similarities distribution of negative pairs and positive pairs on the SOP, In-Shop, CUB and Cars test datasets for the MS method \cite{CVPR-WangHHDS19} and our CBML method based on the ResNet-50 backbone. The first row is the results of the MS method, the second row is our results of the CBML method.}
	\label{fig:similarity-distribution}
\end{figure}

\textbf{(D) Impact of the metric variance constraint (MVC) on the retrieval performance.} In the following experiments, we evaluate the impact of metric variance constraint (\ref{eq:constraint}) with and without the hard pairs mining. These experiments are performed on the CUB dataset with the ResNet-50 backbone. From Table \ref{tab:ab-losses}, we can see that: (1) The metric variance constraint does have significant contribution in both our method and the MS method. Its contribution in the MS method is 1.9\% for the top-1 recall rate. However, in our CBML method without the hard pairs mining, its contribution is much larger, up to 6.0\% for the top-1 recall rate. It should be pointed out that the performance of MS is much better than the CBML if MVC is not used. This is because CBML often easily overfits the training set by bridging the semantic gap between features metric and class label. (2) The hard pair mining is able to improve the performance of our CBML method by 0.1\% and 0.4\% for Recall@1 and Recall@2 rates, respectively.
In Fig. \ref{fig:similarity-distribution}, we plot the distributions of similarity scores for positive pairs (red) and negative pairs (blue) for all four datasets. The top row shows the distributions of the MS method \cite{CVPR-WangHHDS19}. The bottom row shows the distributions of our CBML method.  We can see that our method is able to aggregate positive pairs and separate negative pairs more effectively.

\begin{table}[t]
		\caption{The impact of hard example mining and constraint $\mathcal{L}_2$ on different losses based on the ResNet-50 backbone for the CUB dataset.}
		\begin{center}
		\resizebox{0.8\linewidth}{!}{
			\begin{tabularx}{6.8cm}{l|cccc}
				\hline 
				 Methods & R@1 & R@2 & R@4 & R@8 \\
				\hline
				\hline
				ProxyNCA \cite{ICCV-Movshovitz-Attias17} & 63.6 & 74.5 & 83.4 & 89.2 \\
				\quad + MVC& \textbf{64.7} & \textbf{75.7} & \textbf{83.7} & \textbf{89.5}\\
				MS \cite{CVPR-WangHHDS19} & 65.3 & 75.7 & 84.4 & 90.4 \\
				\quad + MVC& \textbf{67.2} & \textbf{77.4} & \textbf{85.9} & \textbf{91.5}\\
				\cline{1-5}
				CBML & 63.8 & 74.5 & 83.3 & 89.7 \\
				\quad + MVC & 69.8 & 80.0& 87.1& 92.2\\
				\quad\quad + Hard Mining & \textbf{69.9} & \textbf{80.4}& \textbf{87.2}& \textbf{92.5}\\
				\hline
			\end{tabularx}
			}
		\end{center}
		\label{tab:ab-losses}
	\end{table}
	
\begin{table*}[h]
		\caption{Comparison of retrieval performance on the CUB, Cars and SOP datasets for pseudo-supervised transfer learning with 128-dimensional embeddings on GoogLeNet backbone network.}
		\begin{center}
		\resizebox{0.9\linewidth}{!}{
			\begin{tabularx}{15.4cm}{l|cccc||cccc||ccc}
				\hline 
				 \multirow{2}[1]*{Methods}	&\multicolumn{4}{c||}{\textbf{CUB}} &\multicolumn{4}{c||}{\textbf{Cars}}&\multicolumn{3}{c}{\textbf{SOP}}\\
				\cline{2-12} 
				& R@1 & R@2 & R@4 & R@8 & R@1 & R@2 & R@4 & R@8& R@1 & R@10 & R@100\\
				\hline
				\hline
				Examplar [\textcolor{blue}{\footnotesize{TPAMI16}}] \cite{PAMI-DosovitskiyFSRB16} & 38.2 & 50.3 & 62.8 & 75.0 & 36.5 & 48.1 & 59.2& 71.0 & 45.0 & 60.3 & 75.2\\
				NCE [\textcolor{blue}{\footnotesize{CVPR18}}] \cite{CVPR-WuXYL18} & 39.2 & 51.4 & 63.7 & 75.8 & 37.5 & 48.7 & 59.8 & 71.5& 46.6 & 62.3 & 76.8 \\
				DeepCluster [\textcolor{blue}{\footnotesize{ECCV18}}] \cite{ECCV-CaronBJD18} & 42.9 & 54.1 & 65.6 & 76.2 & 32.6 & 43.8 & 57.0 & 69.5 & 34.6& 52.6& 66.8\\
				MOM [\textcolor{blue}{\footnotesize{CVPR18}}] \cite{CVPR-IscenTAC18}& 45.3 & 57.8 & 68.6& 78.4 & 35.5 & 48.2 & 60.6 & 72.4 & 43.3 & 57.2& 73.2\\
				AND [\textcolor{blue}{\footnotesize{ICML19}}] \cite{ICML-Huang19} & 47.3 & 59.4 & 71.0 & 80.0 & 38.4 & 49.6 & 60.2 & 72.9 & 47.4& 62.6 & 77.1\\
				ISIF [\textcolor{blue}{\footnotesize{CVPR19}}] \cite{CVPR-YeZYC19} & 46.2 & 59.0 & 70.1 & 80.2& 41.3 & 52.3 & 63.6 & 74.9& 48.9 & 64.0 & 78.0\\
				sSUML [\textcolor{blue}{\footnotesize{AAAI20}}] \cite{AAAI-DuttaHS20} & 43.5 & 56.2 & 68.3 & 79.1 & 42.0 & 54.3 & 66.0 & 77.2 & 47.8 & 63.6 & 78.3 \\
				aISIF [\textcolor{blue}{\footnotesize{TPAMI20}}] \cite{PAMI-Ye20} & 47.7 & 59.9 & 71.2 & 81.4 & 41.2 & 52.6 & 63.8 & 75.1 & 49.7 & 65.4 & 79.5\\
				CBSwR [\textcolor{blue}{\footnotesize{BMVC20}}] \cite{ArXiv/abs-2009-04091} &47.5& 59.6& 70.6 &80.5& 42.6 & 54.4& 65.4 &76.0&-&-&-\\
				Ortho [\textcolor{blue}{\footnotesize{TAI20}}] \cite{ArXiv-abs-2008-09880} & 47.1 & 59.7 & 72.1 & 82.8 & 45.0 & 56.2 & 66.7 & 76.6 & 45.5 & 61.6 & 77.1\\
			    PSLR [\textcolor{blue}{\footnotesize{CVPR20}}] \cite{CVPR-YeS20} & 48.1 & 60.1 & 71.8 & 81.6 & 43.7 & 54.8 & 66.1 & 76.2 & 51.1 & 66.5 & 79.8\\
				\hline\hline
				\textbf{Ours}: CBML &  \textbf{56.4} & \textbf{68.4} & \textbf{78.6} & \textbf{86.2} & \textbf{45.1}& \textbf{57.1} & \textbf{68.5}& \textbf{78.8}& \textbf{57.9}& \textbf{72.8} & \textbf{84.9}\\
				\textbf{Gain}: CBML & \textcolor{red}{\textbf{+8.3}} & \textcolor{red}{\textbf{+8.3}}& \textcolor{red}{\textbf{+6.5}}& \textcolor{red}{\textbf{+3.4}}& \textcolor{red}{\textbf{+0.1}}& \textcolor{red}{\textbf{+0.9}}& \textcolor{red}{\textbf{+1.8}} & \textcolor{red}{\textbf{+2.2}}&\textcolor{red}{\textbf{+6.8}} & \textcolor{red}{\textbf{+6.3}} & \textcolor{red}{\textbf{+5.1}}\\
				\hline
			\end{tabularx}
			}
		\end{center}
		\label{tab:tl1}
	\end{table*}

\begin{figure}[t]
    \begin{center}
    \includegraphics[width=0.7\linewidth]{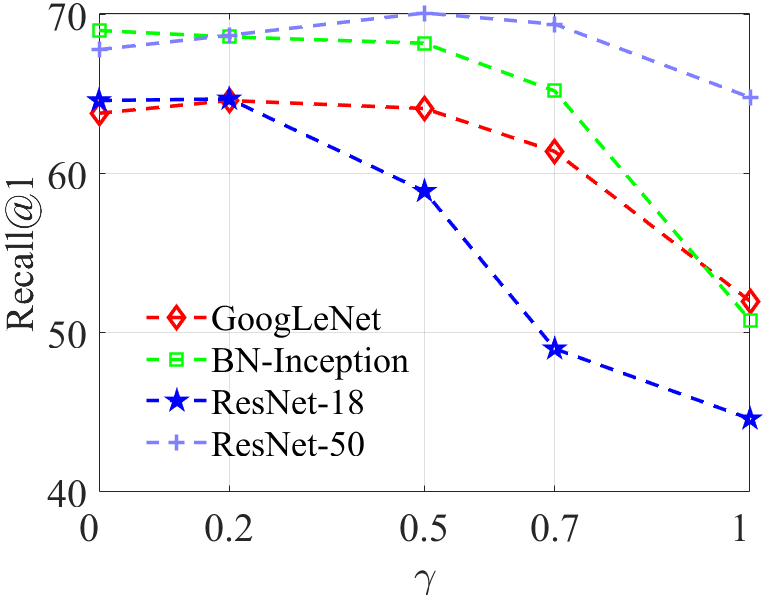}
    \end{center}
    \vspace{-4mm}
	\caption{The impact of the hyperparameter $\gamma$ for different backbones with the CBML method on the CUB dataset.}
	\vspace{-4mm}
	\label{fig:supervised-gamma}
\end{figure}

\textbf{(E) Impact of the hyperparameter $\gamma$ on the CBML performance.} The hyperparameter $\gamma$ in (\ref{eq:hyperplane}) is an important factor for the final performance. Here, we evaluate the impact of this hyperparameter on the CUB dataset with the GoogLeNet, BN-Inception, ResNet-18 and ResNet-50 backbones. Results are shown in Fig. \ref{fig:supervised-gamma}. We can see that the best choice for $\gamma$ are 0.2, 0, 0.2 and 0.5 for GoogLeNet, BN-Inception, ResNet-18 and ResNet-50 backbones, respectively. In our experiments, we set the value of $\gamma$ as 0.2 for all the backbone networks, respectively.

\textbf{(F) Analysis the settings of other algorithm parameters.} Besides the above discussed hyperparameters, we have other parameters in our proposed algorithm, including $\delta^P$ and $\delta^N$, $\alpha^P$, $\beta^P$, $\alpha^N$ and $\beta^N$. Here, we discuss how their values can be chosen appropriately. In the loss function in (\ref{eq:CBML}), $\delta^P$ and $\delta^N$ are constants, which can be incorporated into the exponent by converting $\delta^P=e^{\log\delta^P}$ and $\delta^N=e^{\log\delta^N}$. For the values of $\alpha^P$, $\beta^P$, $\alpha^N$ and $\beta^N$, we can first set the values of $\alpha^P$ and $\alpha^N$ to be $\frac{\mu_P+\mu_N}{2}\approx0.5$ according to (\ref{eq:linear}) and (\ref{eq:log-bayes-transform}). Then, $\frac{\alpha^P}{\beta^P}$ can be set as $\frac{\mu_P^2-\mu_N^2}{2\sigma_P^2}+\log(\delta^P)$, $\frac{\alpha^N}{\beta^N}$ can be set as $\frac{\mu_P^2-\mu_N^2}{2\sigma_N^2}-\log(\delta^N)$. Because the value of $\delta^N=\frac{|\mathcal{P}_i|}{|\mathcal{N}_i|^2}$ is much smaller than the value of $\delta^P=\frac{|\mathcal{N}_i|}{|\mathcal{P}_i|^2}$, thus the value of $\frac{\alpha^P}{\beta^P}$ can be set to a relatively small value and the value of $\frac{\alpha^N}{\beta^N}$ can be set to a relatively large value. This suggests that we can set the initial values of $\beta^P$ and $\beta^N$ as 0.5 and 0.005, respectively. After carefully tuning these parameters for specific distributions, we find that more stable performance can be obtained by setting $\alpha^P=0.5$, $\beta^P=0.5$, $\alpha^N=1.0$ and $\beta^N=0.01$ on the CUB and Cars datasets, and setting $\alpha^P=0.5$, $\beta^P=0.25$, $\alpha^N=0.5$ and $\beta^N=0.05$ on the SOP and In-Shop datasets. Thus, we use these settings in all  experiments.

\subsection{Zero-Shot Image Retrieval on the ImageNet Dataset}
\label{se:zero-shot retrieval}

In the following, we conduct experiments on the large scale ImageNet dataset with zero-shot settings to show the generalization ability of our method. Different from the fine-grained images in CUB, Cars, SOP and In-Shop datasets, images in ImageNet dataset are more general for evaluating zero-shot learning algorithm. We set the batch size as 100 and train the model from scratch. Based on the GoogLeNet backbone, the top-$K$ recall rates of the MS method and our CBML algorithm are shown in Table \ref{tab:imagenet}. The hyper-parameter settings of the MS method are consistent with the MS paper \cite{CVPR-WangHHDS19}. We can see that the proposed CBML method achieves much higher top-$K$ recall rates, improved 4.0\% and 2.4\% for top-1 and top-10 recall rates, respectively.
\begin{table}[h]
		\caption{Top-K recall rates (\%) of the MS and CBML methods with the GoogLeNet backbone on the ImageNet dataset.}
		\begin{center}
		\resizebox{0.8\linewidth}{!}{
			\begin{tabularx}{7.0cm}{l|cc|cc}
				\hline 
				 R@K & \multicolumn{2}{c|}{R@1} & \multicolumn{2}{c}{R@10} \\
				 \hline
			     Methods &MS \cite{CVPR-WangHHDS19} & CBML & MS \cite{CVPR-WangHHDS19} & CBML\\
				\hline
			     Recall Rates & 5.9 & \textbf{9.9} & 22.4 & \textbf{24.8} \\
				\hline
			\end{tabularx}
			}
		\end{center}
		\vspace{-4mm}
		\label{tab:imagenet}
	\end{table}

\subsection{Application to Pseudo-supervised Metric Learning}
\label{se:unsupervised}

In order to verify the generalization capability of our method, we also conduct experiments for pseudo-supervised learning without ground truth labels on the target dataset.
Specifically, we use the $K$-means clustering algorithm to cluster the features  of the training data to obtain the pseudo-label. The total number of clusters  is set as 100 for the CUB and Cars datasets, and 10000 for the SOP dataset.  The results  with the GoogLeNet backbone for 128 dimensional embdddings on the CUB, Cars and SOP datasets are summarized in Table \ref{tab:tl1}. It should be pointed out that these results are obtained using only our loss function (\ref{eq:CBML}) with the generated pseudo-labels. 
We consider two scenarios for performance comparisons: (1) learning with an ImageNet pre-trained model, and (2) learning from scratch.

\textbf{(A) Learning from the ImageNet pre-trained model.} 
In this scenario, we use the network model pre-trained on the ImageNet as the initial backbone encoder and then fine-tune it on the training dataset without using the labels. 
The results  with the GoogLeNet backbone for 128 dimensional embedding on the CUB, Cars, and SOP datasets are summarized in Table \ref{tab:tl1}. These results are obtained using our proposed CBML loss function (\ref{eq:CBML}) with clustering-based iterative pseudo label generation.

From Table \ref{tab:tl1}, we can see that our CBML method outperforms the state-of-the-art methods by large margins. Our  method has improved the Recall@1, Recall@2, Recall@4 and Recall@8 rates by 8.3\%, 8.3\%, 6.5\%, 3.4\%, respectively, on the CUB dataset, and 0.1\%, 0.9\%, 1.8\%, 2.2\% on the Cars dataset. On the SOP dataset, our method has improved the Recall@1, Recall@10, and Recall@ 100 rates by 6.8\%, 6.3\%, and 5.1\%, respectively.

\textbf{(B) Learning from scratch.} Following the aISIF \cite{PAMI-Ye20} method, we also test the performance using a randomly initialized ResNet-18 network without pre-training, on the large-scale SOP dataset, as shown in Table \ref{tab:ab-initialize}. Results demonstrate that the proposed method achieves much better performance than other methods, 7.7\%, 7.1\%, and 5.8\% gain over the PSLR method for Recall@1, Recall@10 and Recall@100 rates. 

\begin{table}[t]
		\caption{Comparisons of retrieval performance on the SOP dataset with 128-dimensional embeddings on the ResNet-18 backbone network without pre-trained parameters.}
 		\vspace{-2mm}
		\begin{center}
		\resizebox{0.7\linewidth}{!}{
			\begin{tabularx}{5.5cm}{l|ccc}
				\hline 
				 Methods&  R@1 & R@10 & R@100\\
				\hline
				\hline
				Examplar \cite{PAMI-DosovitskiyFSRB16} & 31.5 & 46.7 & 64.2\\
				NCE \cite{CVPR-WuXYL18} & 34.4 & 49.0 & 65.2 \\
				MOM \cite{CVPR-IscenTAC18}& 16.3 & 27.6 & 44.5\\
				AND \cite{ICML-Huang19} & 36.4& 52.8 & 67.2\\
				ISIF \cite{CVPR-YeZYC19} & 39.7 & 54.9 & 71.0\\
				aISIF \cite{PAMI-Ye20} & 40.7 & 55.9 & 72.2\\
				PSLR \cite{CVPR-YeS20} & 42.3 & 57.7 & 72.5 \\
				\hline\hline
				\textbf{Ours}: CBML &  \textbf{50.0} & \textbf{64.8} & \textbf{78.3}\\
				\textbf{Gain}: CBML & \textcolor{red}{\textbf{+7.7}}& \textcolor{red}{\textbf{+7.1}} & \textcolor{red}{\textbf{+5.8}}\\
				\hline
			\end{tabularx}
			}
		\end{center}
 		\vspace{-4mm}
		\label{tab:ab-initialize}
	\end{table}
	
\textbf{(C) Learning with different backbone networks.} Following the aISIF \cite{PAMI-Ye20} method,  we also conduct experiments with the ResNet-18 and ResNet-50 backbone encoders for our CBML method. The embedding size is set to be 128.
Results of top-1 recall rates on the CUB, Cars and SOP datasets are shown in Table \ref{tab:ab-backbone}. Our proposed CBML method benefits from stronger backbone encoders and outperforms the existing method. It should be noted that we could only provide comparison with the aISIF \cite{PAMI-Ye20} and the PSLR \cite{CVPR-YeS20} papers since other papers did not report results on other backbone networks.

\begin{table}[t]
		\caption{Top-1 recall rates (\%) with 128-dimensional embeddings on different backbone networks.}
		\vspace{-2mm}
		\begin{center}
		\resizebox{0.8\linewidth}{!}{
			\begin{tabularx}{6.3cm}{l|c|ccc}
				\hline 
				Backbone &Methods & CUB & Cars & SOP\\
				\hline
				\hline
				\multirow{3}[1]*{GoogLeNet} & aISIF \cite{PAMI-Ye20}&  47.7 & 41.2 & 49.7\\
				& PSLR \cite{CVPR-YeS20} & 48.1 & 43.7 & 51.1 \\
				& CBML & \textbf{56.4} & \textbf{45.1} & \textbf{57.9}\\
				\hline
				\multirow{3}[1]*{ResNet-18} & aISIF \cite{PAMI-Ye20}& 45.5 & 34.9 & 54.7\\
				 & PSLR \cite{CVPR-YeS20} & 48.9 & 39.2 & 52.2 \\
				& CBML & \textbf{51.7} & \textbf{39.4} & \textbf{57.1}\\
				\hline
				\multirow{3}[1]*{ResNet-50} & aISIF \cite{PAMI-Ye20}& 47.3 & 41.4 & 55.6 \\
				& PSLR \cite{CVPR-YeS20} & 49.0 & 42.8 & \textbf{61.6}\\
				& CBML & \textbf{59.2}& \textbf{48.8} & 59.5\\
				\hline
			\end{tabularx}
			}
		\end{center}
		\vspace{-4mm}
		\label{tab:ab-backbone}
	\end{table}

\textbf{(D) Ablation experiments for pseudo-supervised learning.} Table \ref{tab:ab-constraint-hard} summarizes the contributions of major components of our algorithm, namely, contrastive Bayesian loss and the metric variance constraint in the pseudo-supervised metric learning setting based on the GoogLeNet backbone. The following observation can be made from the comparison results in Table \ref{tab:ab-constraint-hard}. (1) The metric variance constraint does have significant contribution in our method for unsupervised metric learning. Its contribution in our CBML method is about 2.9\% for the top-1 recall rate. (2) The hard pair mining cannot further boost the performance of our pseudo-supervised metric learning based on CBML method. Thus, we do not use the hard pair mining technique in our pseudo-supervised experiments.
From the above performance comparisons for pseudo-supervised metric learning, we can see that our proposed constrastive Bayesian analysis is generalizable, being able to achieve significant performance gain on a wide range of metric learning applications.

\begin{table}[t]
		\caption{The contributions of the major components of our CBML method using GoogLeNet backbone on the CUB dataset.}
		\begin{center}
		\resizebox{\linewidth}{!}{
			\begin{tabularx}{8.0cm}{l|cccc}
				\hline 
				 Methods & R@1 & R@2 & R@4 & R@8 \\
				 \hline
				 PSLR (current state of the art)  & 48.1 & 60.1 & 71.8 & 81.6\\
				\hline
			    CBML & 53.5 & 65.5 & 76.2& 84.8\\
			    \quad + MVC & \textbf{56.4} & \textbf{68.4} & \textbf{78.6} & \textbf{86.2}\\
			    \quad\quad + Hard Mining & 56.3 & 68.2& 78.6& 86.0\\
				\hline
			\end{tabularx}
			}
		\end{center}
		\vspace{-4mm}
		\label{tab:ab-constraint-hard}
	\end{table}
	
Table \ref{tab:ab-batch-size} presents the impact of different batch sizes on the CUB dataset with the ResNet-18 backbone.  From Table \ref{tab:ab-batch-size}, we can see that larger batch sizes can lead to better performance, and the performance  saturates when the batch size is between 80 and 200. In our experiments, we random set the batch size between 80 and 200 in each epoch for our pseudo-supervised learning. 

\begin{table}[h]
		\caption{Top-1 recall rates (\%) of different batch size with the ResNet-18 backbone on the CUB dataset.}
		\begin{center}
		\resizebox{0.7\linewidth}{!}{
			\begin{tabularx}{5.3cm}{l|cccc}
				\hline 
				 Batch Size & 30 & 80 & 100 & 200\\
				\hline
				\hline
				Recall@1 & 48.9 & 51.4 & 51.7 & 51.7\\
				\hline
			\end{tabularx}
			}
		\end{center}
		\vspace{-4mm}
		\label{tab:ab-batch-size}
	\end{table}

\subsection{Further Discussion}
From the above experiments, we have the following observations. 

\textbf{(1) About the training batch sizes}. By conducting experiments on the SOP and In-Shop datasets with large batch sizes, Wang \textit{et al.} \cite{CVPR-WangHHDS19} demonstrated that training with large batch sizes can significantly improve the feature embedding performance. Then, they proposed a memory-bank \cite{CVPR-Wang20} technique to reduce the memory consumption on the GPU and introduced hard example mining over the whole training set. However, only a small batch size is needed (Table \ref{tab:ab-batch-size}) in our work to obtain the state-of-the-art feature embedding performance, and we have verified that the retrieval performance of our method cannot be further improved by the memory-bank technique. 

\textbf{(2) About hard example mining}. From Table \ref{tab:ab-losses} and Table \ref{tab:ab-constraint-hard}, we can see that the performance improvement by hard example mining is very limited in our method when metric variance constraint (MVC) is used. Although various methods have been developed in the literature to improve the performance of hard example mining, it is not necessary when MVC is used.

\textbf{(3) About metric variance constraint for generalization ability}.  According to Table \ref{tab:ab-distribution-constraint} and Table \ref{tab:ab-losses}, we can see that metric variance constraint is very important for training a generalizable model. Besides this, the MVC can be used to verify whether better performance can be obtained by overfitting the training set for a dataset. 

\textbf{(4) About the major difference between our method and existing deep metric learning methods}. Existing metric learning methods aim to enforce the distance between similar and dissimilar samples directly, which optimizes the embedding performance locally. However, our CBML method directly optimizes the global distribution of metric distance to achieve improved performance.

\section{Conclusion}
In this paper, we have developed a contrastive Bayesian analysis to bridge the semantic gap between features at intermediate feature layers and class label decision at the final output layer. Based on this analysis, we are able to model and predict the posterior probabilities of image labels conditioned by their features similarity in a contrastive learning setting. This contrastive Bayesian analysis leads to a new loss function for deep metric learning. To improve the generalization capability of the proposed method onto new classes, we further extend the contrastive Bayesian loss with a metric variance constraint.
Our experimental results and ablation studies have demonstrated that the CBML method has significantly improved the performance of deep metric learning, outperforming existing methods by a large margin. This new contrastive Bayesian analysis can be directly applied or further extended for a set of other learning applications. 

\section*{Acknowledgments}
This work was supported in part by the National Key R\&D Program of China 2021YFE0110500, in part by the National Natural Science Foundation of China under Grant 62202499, 61872034, 62062021, and 62011530042, in part by the Hunan Provincial Natural Science Foundation of China under Grant 2022JJ40632, in part by the Beijing Municipal Natural Science Foundation under Grant 4202055, in part by the Natural Science Foundation of Guizhou Province under Grant [2019]1064.

\bibliographystyle{IEEEtran}
\bibliography{egbib}

\begin{IEEEbiography}[{\includegraphics[width=1in,height=1.25in,clip,keepaspectratio]{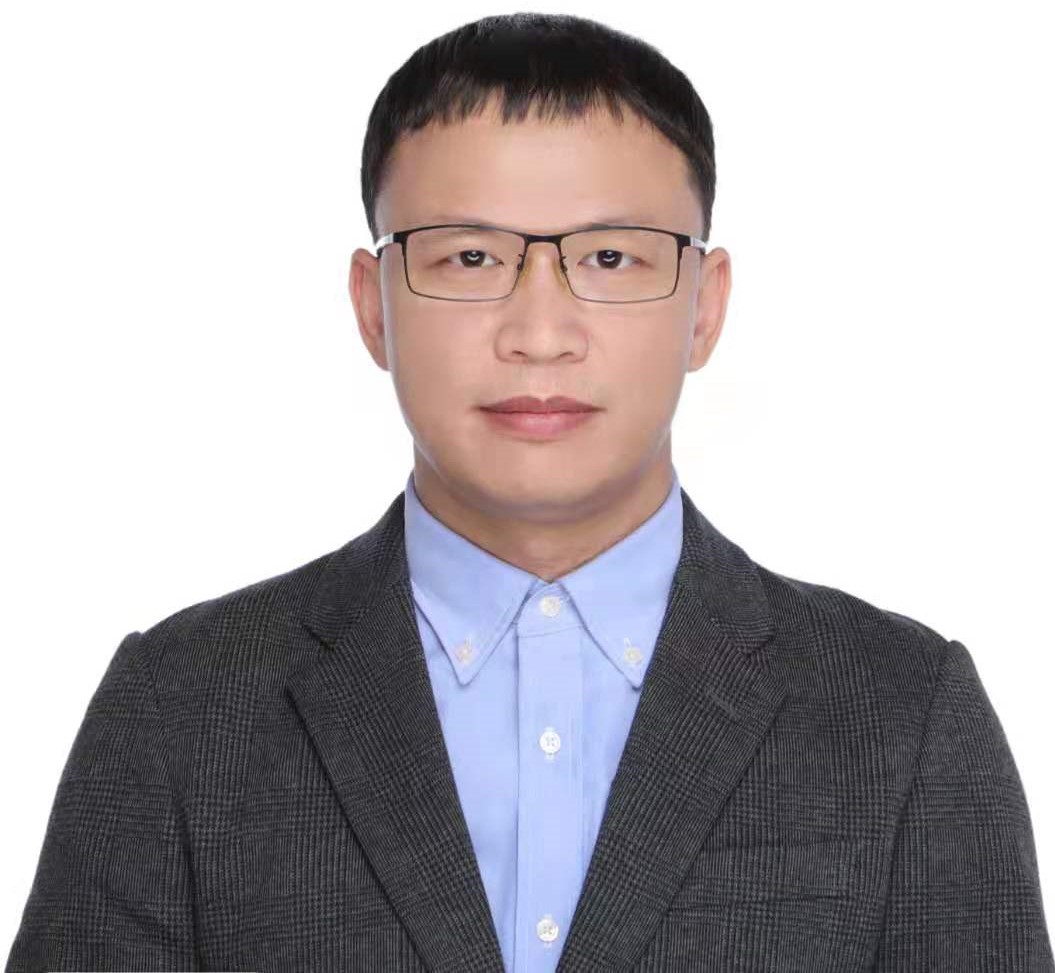}}]{Shichao Kan}
received the B.E., M.S. and Ph.D. degrees from the School of Computer and Information Science, Beijing Jiaotong University, Beijing, China, in 2014, 2016 and 2021, respectively. From 2019 to 2020, he was a visiting student researcher with the Department of Computer Science, University of Missouri, Columbia, MO, USA. He is currently a lecturer with the School of Computer Science and Engineering, Central South University, Hunan, China. His research interests include metric learning, large-scale image retrieval, object search, and deep learning.
\end{IEEEbiography}

\begin{IEEEbiography}[{\includegraphics[width=1in,height=1.25in,clip,keepaspectratio]{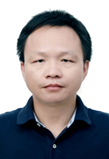}}]{Zhiquan He}
is currently an Assistant Professor in College of Information Engineering, Shenzhen University, China.  He is the Associate Director of Multimedia Information Service Engineering
Technology Research Center. He received his M.S. degree from Institute of Electronics, Chinese Academy of Sciences in 2001, and the PhD degree from the department of Computer Science, University of Missouri-Columbia in 2014. He worked at Mathworks as a research engineer. His research areas include image processing, computer vision, and machine learning.
\end{IEEEbiography}

\begin{IEEEbiography}[{\includegraphics[width=1in,height=1.25in,clip,keepaspectratio]{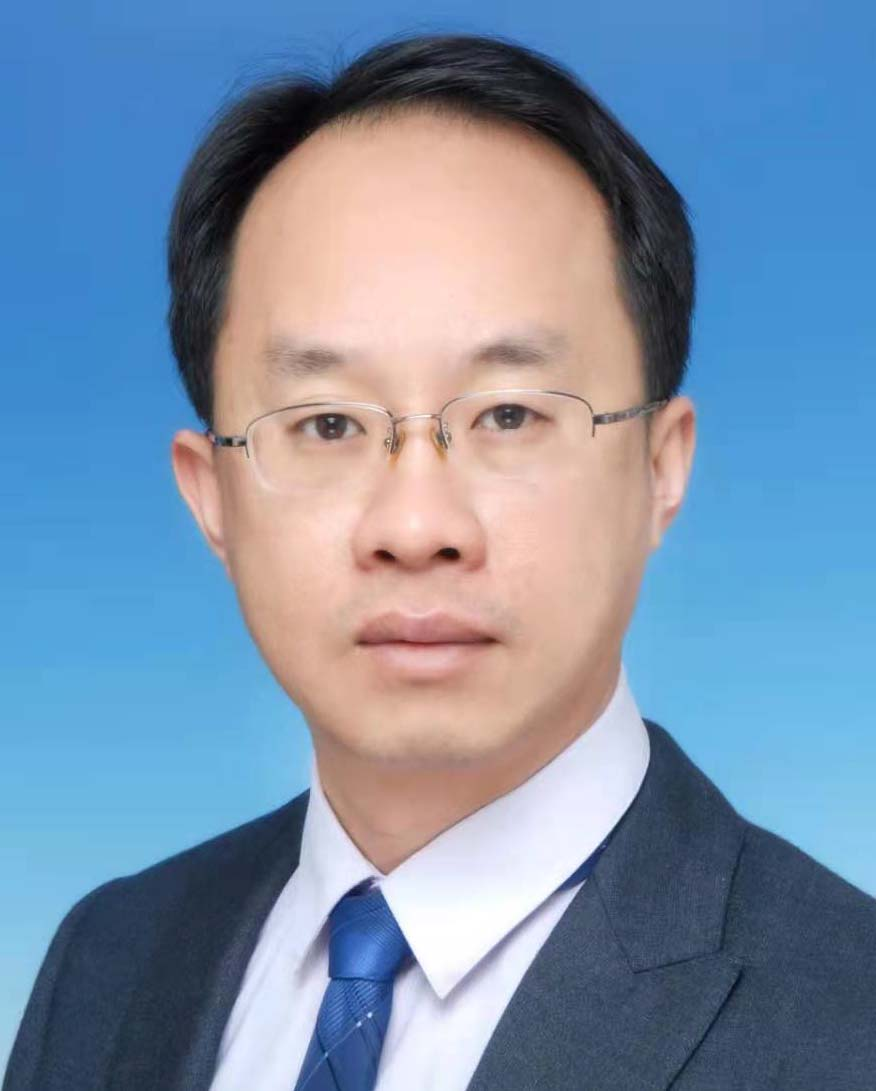}}]{Yigang Cen}
received the Ph.D. degree in control science engineering from the Huazhong University of Science Technology, Wuhan, China, in 2006. In 2006, he joined the Signal Processing Centre, School of Electrical and Electronic Engineering, Nanyang Technological University, Singapore, as a Research Fellow. From 2014 to 2015, he was a Visiting Scholar with the Department of Computer Science, University of Missouri, Columbia, MO, USA. He is currently a Professor and a Supervisor of doctoral students with the School of Computer and Information Technology, Beijing Jiaotong University, Beijing, China. His research interests include computer vision, intelligent transportation and intelligent security, etc.
\end{IEEEbiography}

\begin{IEEEbiography}[{\includegraphics[width=1in,height=1.25in,clip,keepaspectratio]{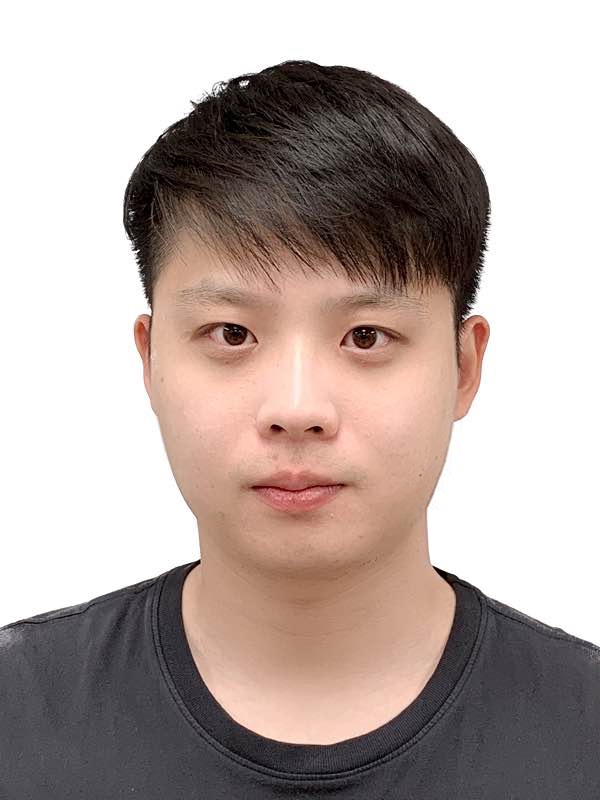}}]{Yang Li}
is currently working toward the Ph.D. degree with the Department of Electrical Engineering and Computer Science, University of Missouri, Columbia, MO, USA. His current research interests include semi-supervised learning, unsupervised learning, video compression, and metric learning.
\end{IEEEbiography}

\begin{IEEEbiography}[{\includegraphics[width=1in,height=1.25in,clip,keepaspectratio]{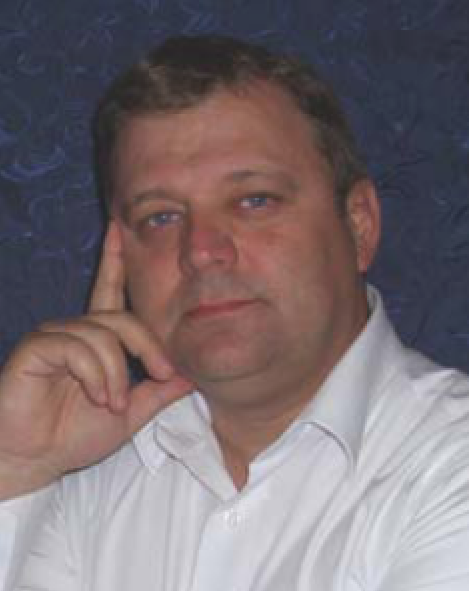}}]{Vladimir Mladenovic}
is currently an associate professor with the Faculty of Technical Sciences Cacak University of Kragujevac. His research interests include wireless communication and image processing.
\end{IEEEbiography}

\begin{IEEEbiography}[{\includegraphics[width=1in,height=1.25in,clip,keepaspectratio]{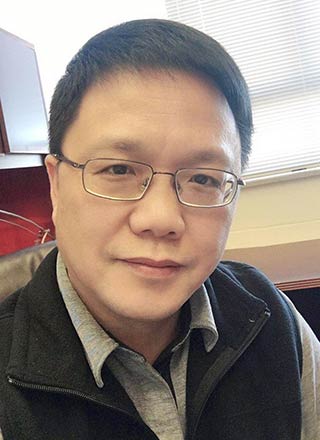}}]{Zhihai He} (IEEE Fellow 2015) received the B.S.degree in mathematics from Beijing Normal University, Beijing, China, in 1994, the M.S. degree in mathematics from the Institute of Computational Mathematics, Chinese Academy of Sciences, Beijing, China, in 1997, and the Ph.D. degree in electrical engineering from the University of California, at Santa Barbara, Santa Barbara, CA, USA, in 2001. In 2001, he joined Sarnoff Corporation, Princeton, NJ, USA, as a member of technical staff. In 2003, he joined the Department of Electrical and Computer Engineering, University of Missouri, Columbia MO, USA, where he was a Tenured Full Professor. He is currently a chair professor with the Department of Electrical and Electronic Engineering, Southern University of Science and Technology, Shenzhen, P. R. China. His current research interests include image/video processing and compression, wireless sensor network, computer vision, and cyber-physical systems.

He is a member of the Visual Signal Processing and Communication Technical Committee of the IEEE Circuits and Systems Society. He serves as a technical program committee member or a session chair of a number of international conferences. He was a recipient of the 2002 IEEE TRANSACTIONS ON CIRCUITS AND SYSTEMS FOR VIDEO TECHNOLOGY Best Paper Award and the SPIE VCIP Young Investigator Award in 2004. He was the Co-Chair of the 2007 International Symposium on Multimedia Over Wireless in Hawaii. He has served as an Associate Editor for the IEEE TRANSACTIONS ON CIRCUITS AND SYSTEMS FOR VIDEO TECHNOLOGY (TCSVT), the IEEE TRANSACTIONS ON MULTIMEDIA (TMM), and the Journal of Visual Communication and Image Representation. He was also the Guest Editor for the IEEE TCSVT Special Issue on Video Surveillance.
\end{IEEEbiography}

\end{document}